\documentclass[lettersize,journal]{IEEEtran}
\usepackage{amsmath,amsfonts}
\usepackage{algorithmic}
\usepackage{array}
\usepackage{multirow}
\usepackage{textcomp}
\usepackage{stfloats}
\usepackage{url}
\usepackage{color}
\usepackage{times}
\usepackage{verbatim}
\usepackage{graphicx}
\usepackage{hyperref}
\usepackage{amssymb}
\usepackage{subfigure}
\usepackage{amsmath} 
\usepackage{booktabs}
\usepackage{cite}

\hypersetup{hypertex=true,
	colorlinks=true,
	linkcolor=blue,
	anchorcolor=blue,
	citecolor=blue}
\begin{document}
\title{Modelling Power Consumptions for Multi-rotor UAVs}
\author{Hao Gong, Baoqi Huang, Bing Jia and Hansu Dai
\thanks{Hao Gong, Baoqi Huang and Bing Jia are with the College of Computer Science, Inner Mongolia University, Hohhot 010021, China (e-mail: gonghao@mail.imu.edu.cn; cshbq@imu.edu.cn; jiabing@imu.edu.cn).

Hansu Dai is with the Chengdu Aircraft Design and Research institute, Chengdu 610041, China (e-mail: captaindai2004@163.com).}
}

\maketitle
\begin{abstract}
Unmanned aerial vehicles (UAVs) have various advantages, but their practical applications are influenced by their limited energy. Therefore, it is important to manage their power consumptions and also important to establish corresponding power consumption models. However, most of existing works either establish theoretical power consumption models for fixed-wing UAVs and single-rotor UAVs, or provide heuristic power consumption models for multi-rotor UAVs without rigorous mathematical derivations. This paper aims to establish theoretical power consumption models for multi-rotor UAVs.
To be specific, the closed-form power consumption models for a multi-rotor UAV in three flight statuses, i.e., forward flight, vertical ascent and vertical descent, are derived by leveraging the relationship between single-rotor UAVs and multi-rotor UAVs in terms of power consumptions. On this basis, a generic flight power consumption model for the UAV in a three-dimensional (3-D) scenario is obtained. Extensive experiments are conducted by using DJI M210 and a mobile app made by DJI Mobile SDK in real scenarios, and confirm the correctness and effectiveness of these models; in addition, simulations are performed to further investigate the effect of the rotor numbers on the power consumption for the UAV. The proposed power consumption models not only reveal how the power consumption of multi-rotor UAVs are affected by various factors, but also pave the way for introducing other novel applications. 
\end{abstract}


\section{Introduction}
\IEEEPARstart{N}{owadays}, unmanned aerial vehicles (UAVs) have found a significant number of applications in wireless communication and power grid inspection, etc., due to their decreasing expense and increasing functionality \cite{zeng2016wireless,agrawal2022finite,pham2022energy,zhang2021energy,jenssen2019intelligent}. However, energy limitation is always a key and unavoidable issue in UAV applications \cite{qi2022energy,yang2020energy}. which leads to more studies focusing on the power consumption for the UAVs.
Additionally, UAV power consumption consists of two main parts: one is the conventional communication-related power consumption due to, e.g., signal processing, circuits and power amplification; the other is the flight power consumption to ensure that the UAV remains aloft and supports its movement. In general, the flight power consumption depends on the flight status and is usually much more significant than communication related power expenditure. Thus, it is important to understand how the UAV power consumption varies with its flight status and also important to model the flight power consumption.

Prior researches on the UAV power consumption models can be loosely classified into three categories. The first category is simply applying the existing power models of the ground vehicles or robots to UAVs \cite{han2015traffic,tseng2017personalized,sun2017latency,thibbotuwawa2018energy,juan2021energy}. Whereas, the power consumption for the mobile vehicles or robots moving on the ground are not suitable for UAVs due to their fundamentally different moving mechanisms. Therefore, more effective and feasible models as the second category are proposed, i.e., the experimental-driven heuristic power consumption models based on measured/simulated data \cite{ahmed2016energy,di2015energy,tseng2017flight,chan2020procedure,abeywickrama2018comprehensive}. For instance, in \cite{ahmed2016energy}, some brief experiments have been performedand to state the power consumption for a few basic UAV (quadrotor UAV) maneuvering actions. Similarly, a power consumption model for a specific UAV (quadrotor UAV) is given in cite \cite{di2015energy} based on experiment results, which involves its speed and operating conditions. However, both \cite{ahmed2016energy} and \cite{di2015energy} were simply expressed in relation to speed and do not contain enough information for comparisons, the lack of closed-form expression limits their applications. Besides, by analyzing the battery performance, some power consumption models for electricpowered UAVs were obtained in \cite{tseng2017flight,chan2020procedure,abeywickrama2018comprehensive}. In order to further understand the power consumption for the UAV theoretically and improve the applicability of the models, the last category  considers the theoretical-driven power consumption model by kinematic and aircraft theory \cite{filippone2006flight,johnson2012helicopter,bramwell2001bramwell}, such as \cite{zeng2017energy,zeng2019energy,yan2021new,yang2019energy,gao2021energy,liu2017power}, for example, a generic power consumption model as a function for the UAV's velocity and acceleration was derived for fixed-wing UAVs in \cite{zeng2017energy}, whereas, the above results for fixed-wing UAVs cannot be applied for rotary-wing UAVs, due to their fundamentally different mechanical designs and hence drastically different power consumption models. Furthermore, in \cite{zeng2019energy}, the authors derived a closed-form power consumption model for single-rotor UAVs in a one-dimensional (1-D) forward flight at a constant speed without acceleration/deceleration. Reference \cite{shan2020looking,babu2021cost} have used the power consumption model in \cite{zeng2019energy} to study the energy-efficient UAV communications. 
Considering the practical situation, the authors in \cite{yan2021new,yang2019energy} extended the result in \cite{zeng2019energy} by deriving an analytical model for single-rotor UAVs in a two-dimensional (2-D) forward flight, besides, the power consumption for single-rotor UAVs in vertical flight is also involved in \cite{yang2019energy}. In addition, a validation of the theoretical model by measured data was conducted in \cite{gao2021energy} to ensure the correctness of the model unlike \cite{liu2017power} which only aims to obtain power model expressions by fitting measurement data.
All the above works have provided very valuable guidance on establishing the power consumption models for the UAVs.

However, considering the power consumption models for multi-rotor UAVs which are the most popular kind of UAVs today, existing works \cite{ahmed2016energy,di2015energy,tseng2017flight,chan2020procedure} only gave the heuristic power consumption models without rigorous mathematical derivation, limiting their applications in many research fields (e.g., design for UAV communications, path planning for UAVs, etc.). Therefore, great efforts have to be devoted to deriving theoretical power consumption models for multi-rotor UAVs to satisify the needs. 
Additionally, the prior studies \cite{zeng2019energy,yan2021new,yang2019energy} on the theoretical power consumption models only considered the UAVs' flight statuses in a 1-D or 2-D scenarios, while leaving a more general flight scenario, i.e., three-dimensional (3-D) scenario unaddressed. This thus motivate more works to study the corresponding power consumption characteristics for the UAVs in a 3-D scenario.

This paper tackles the problem of lacking theoretical power consumption models for multi-rotor UAVs in general flight scenarios. To be specific, 
first, the power consumption models with closed-form expression for a multi-rotor UAV in horizontal flight and vertical flight are established based on the results in \cite{zeng2019energy}, \cite{yang2019energy}, enabling us to theoretically investigate the factors affecting to the required power. Then, a general flight power consumption model for the UAVs in a 3-D scenario is extended by intergrating the models derived above. Afterwards, for the purpose of verifying the correctness of the power consumption models, extensive experiments were conducted and more than 7000 valid power-speed data points were collected. Last, simulation was performed to further analyze the effect of the rotor numbers on the power consumption and indicated that the power consumption decreases with the rotor numbers incresing, but the effect becomes smaller with the UAV flight speed increasing.
\section{Single-Rotor UAV Power Consumption Models}
On the one hand, in order to better analyze and utilize the energy of single-rotor UAVs, various power consumption models for single-rotor UAVs are reported. On the other hand, since the multi-rotor UAVs can be regarded to some extent as extension of the single-rotor UAVs, the single-rotor UAV power consumption models can be used as the basis for formulating the power model for multi-rotor UAVs. Therefore, prior to introducing our multi-rotor UAV power consumption model, it is necessary to understand the composition of the power consumption models for single-rotor UAVs. To this end, in this section, we give a brief overview on the theoretical power consumption models for single-rotor UAVs with closed-form expression in the literature. Specifically, we present the analytical models describing the power consumption for a single-rotor UAV in horizontal flight and vertical flight, which are respectively derived in \cite{zeng2019energy} and \cite{yang2019energy}. The two types of models reflect the power consumption for the UAV in different flight statuses, but the physical meaning of the notations are the same. The main notations in  this section are summarized in Table \ref{tabl1}.
\begin{table}[h]
	\centering
	\begin{center}
		\caption{List of Main Notations.}\label{tabl1}
		\begin{tabular}{|c|p{4.1cm}|c|}
			\hline
			\textbf{Notation} & \textbf{Physical meaning} & \textbf{Simulation value} \\
			\hline
			\multirow{1}{*}{$\delta$} & Profile drag coefficient & 0.011\\
			\hline
			\multirow{1}{*}{$\rho$} & Air density in $kg/m^3$ & 1.168\\
			\hline
			\multirow{1}{*}{$s$} & Rotor solidity & 0.045\\
			\hline
			\multirow{1}{*}{$A$} & Rotor disc area in $m^2$ & 0.214\\
			\hline
			\multirow{2}{*}{$\Omega$} & Blade angular velocity in radians/second & \multirow{2}{*}{-}\\
			\hline
			\multirow{1}{*}{$R$} & Rotor radius in meter (m) & 0.26\\
			\hline
			\multirow{2}{*}{$k$} & Incremental correction factor to induced power & \multirow{2}{*}{0.11}\\
			\hline
			\multirow{1}{*}{$W$} & UAV weight in Newton & 20\\
			\hline
			\multirow{1}{*}{$\tilde{\kappa}$} & Thrust-to-weight ratio, $\tilde{\kappa} \triangleq \frac{T}{W}$ & 1\\
			\hline
			\multirow{2}{*}{$v_0$} & Mean rotor induced velocity in hover & \multirow{2}{*}{6.325}\\
			\hline
			\multirow{2}{*}{$S_{FP\parallel}$} & Fuselage equivalent flat plate area in horizontal status in $m^2$ & \multirow{2}{*}{0.009}\\
			\hline
		\end{tabular}
	\end{center}
\end{table}
\subsection{Power Consumption Models for a Single-Rotor UAV in Horizontal Flight}
In this subsection, the required power for a single-rotor UAV flying forward in the horizontal direction is modeled, and in particular, a power consumption model for hovering, which is a situation the flight speed equals to 0 is also considered. 

According to \cite{zeng2019energy}, the power consumed by a single-rotor UAV to hover can be expressed as 
\begin{equation}\label{A}
	P_{1}=\underbrace{\frac{\delta}{8} \rho s A \Omega^{3} R^{3}}_{\triangleq P_{0}}+\underbrace{(1+k) \frac{W^{3 / 2}}{\sqrt{2 \rho A}}}_{\triangleq P_{i}},
\end{equation}
where $P_0$ and $P_{i}$ are two constants representing the \emph{blade profile power} and \emph{induced power}, respectively. Based on \eqref{A}, we obtain the power consumption for hovering status as $P_{1}=P_0+P_{i}$, which is a finite value depending on the UAV weight, air density, rotor disc area, etc. 


Further, provided that the single-rotor UAV is flying forward at a constant speed, say $V$, the corresponding power consumption can be expressed as 
\begin{equation}\label{B}
	\begin{aligned}
		P_2(V, \tilde{\kappa})=& P_{0}\left(1+\frac{3 V^{2}}{\Omega^{2} R^{2}}\right)
		+P_{i} \tilde{\kappa}\left(\sqrt{\tilde{\kappa}^{2}+\frac{V^{4}}{4 v_{0}^{4}}}-\frac{V^{2}}{2 v_{0}^{2}}\right)^{1 / 2}\\
		+&\frac{1}{2}S_{FP\parallel}\rho V^{3},
	\end{aligned}
\end{equation}
where the first two terms in \eqref{B} denote the \emph{blade profile power} and \emph{induced power} in forward flight status, respectively, and are dependent on the specific speed $V$ instead of staying constant as in the hovering status. $S_{FP\parallel}\rho V^3/2$ denotes the \emph{parasite power}. It can be observed from \eqref{B} that the blade profile power and parasite power increase quadratically and cubically with $V$, respectively, and are necessary for overcoming the profile drag of the blades and the fuselage drag \cite[eq.(4.5)]{bramwell2001bramwell}, respectively. Besides, the induced power can be regarded as the power to overcome the induced drag of the blades, which decreases with $V$.

\subsection{Power Consumption Models for a Single-Rotor UAV in Vertical Flight}
Based on the results in \cite{zeng2019energy}, the closed-form expression of single-rotor UAV power consumption model in vertical flight was reported in \cite{yang2019energy}, i.e.

\begin{equation}\label{C}
	P_{3}(V, T)=P_{1}+\frac{1}{2} T V+\frac{T}{2} \sqrt{V^{2}+\frac{2 T}{\rho A}},
\end{equation}
where $T$ is the rotor thrust, and the other parameters are the same as those introduced in the above. The vertical flight of single-rotor UAVs usually include vertical ascent and vertical descent, which should have different values of $T$ because of the corresponding drag directions are different. Hence, with the same constant speed $V$, the corresponding required powers for a single-rotor UAV in vertical ascent and vertical descent are certainly different. Meanwhile, it can be obtained from \eqref{C} that the speed $V$ is assumed nonzero only in the vertical direction. 

In combination with \eqref{B}, it can be concluded that $V$ and $T$ are the two key factors affecting the required power of a single-rotor UAV in flight status.

The above closed-form formula \eqref{A}-\eqref{C} describe the power incurred by a single-rotor UAV in different flying status, but cannot be directly applied to a multi-rotor UAV. To this end, we attempt to establish power consumption models for multi-rotor UAVs in Section \uppercase\expandafter{\romannumeral3} with closed-form expressions, which can obviously be applied in more practical scenarios and applications.

\section{Multi-Rotor UAV Power Consumption Models}
In this section, we first introduce the multi-rotor UAVs considered in the subsequent analysis and an abstract representation for the power consumption of a multi-rotor UAV, then derive closed-form power consumption models for a multi-rotor UAV in horizontal flight and vertical flight both with a constant speed  in 1-D or 2-D scenarios, 
and finally further extend the above models to 3-D scenarios. 

\subsection{Analysis of Multi-Rotor UAVs}
There are many classification methods of multi-rotor UAVs, e.g. multi-rotor UAVs with even rotors and odd rotors according to the parity of the number of rotors. However, the application scope of the former is wider than that of the latter due to its simple flight control mechanism. Therefore, for the purposes of conducting tractable analyses, this paper only deals with multi-rotor UAVs with even rotors, such as quadrotor UAVs, hexarotor UAVs and etc., and shall not consider double-layer multi-rotor UAVs. The extension to general multi-rotor UAVs is left as our future work. 

In order to apply the models in relation to single-rotor UAVs in deriving the power consumption models for multi-rotor UAVs, we have to establish the relationship between multi-rotor UAVs and single-rotor UAVs in terms of power consumptions. Given a considered multi-rotor UAV, it is commonly assumed that every rotor is inentical, and is symmetrically distributed \cite{walid2014modeling}, therefore the axial momentum theory applies \cite{bramwell2001bramwell};
a multi-rotor UAV are assumed to be composed of multiple identical single-rotor UAVs and the number of single-rotor UAVs depends on the number of rotors in a multi-rotor UAV. On this basis, the power consumption for a multi-rotor UAV can be approximately evaluated by summing power consumptions of multiple single-rotor UAVs; in order to facilitate the subsequent analysis and derive the power consumption models, define the following notations:
\begin{itemize}
	\item A multi-rotor UAV of weight $W$ has $n$ rotors;
	\item The weight assigned to each rotor is $W_r$, i.e., $W_r=\frac{W}{n}$;
	\item The thrust of a multi-rotor UAV is defined as $T$;
	\item The thrust of the $i$-th rotor in a multi-rotor UAV is $T_i$, i.e., $T=\sum T_i$;
\end{itemize}

Given a specific multi-rotor UAV, the parameters of each rotor include $\delta$, $A$, $s$, $C_T$, $k$, $v_0$, $\rho$, $S_{FP\parallel}$ and $S_{FP\perp}$, where $C_T$ is the thrust coefficient, $S_{FP\perp}$ is the fuselage equivalent flat plate area in the vertical status and the others are the same as those in Table \ref{tabl1} 

Afterwords, we shall derive the power consumption models for a multi-rotor UAV in both horizontal flight and vertical flight based on the above assumptions. 
\subsection{Power Consumption Models for a Multi-Rotor UAV in Horizontal Flight}
Prior to presenting the models, let us consider $\Omega$ which represents the angular velocity of rotors in \eqref{A}-\eqref{C}. In actual flights, it is known that the flight of a multi-rotor UAV is controled by adjusting the angular velocity of each rotor, such that different angular velocities of different rotors produce different thrust. Since the thrust $T$ is proportional to squared blade angular velocity $\Omega$, i.e., $\Omega=\sqrt{\frac{T}{C_T\rho AR^2}}$ given by \cite[eq.(11.1)]{bramwell2001bramwell}, $\Omega$ is substituted by this equation involving $T$ and other parameters what follows.

In horizontal flight, the thrust provided by each rotor of the multi-rotor UAV differs according to the specific flight status, e.g., hovering and forward flight. When the multi-rotor UAV is hovering, the thrust $T$ balances the UAV's weight, so that the $i$-th rotor should provide the same thrust $T_i$, i.e., $T_i=W_r=\frac{W}{n}$. Based on the power consumption for the single-rotor UAV in hovering status in \eqref{A}, the corresponding required power for the $i$-th rotor in the multi-rotor UAV, denoted $P_{h}^i$, can be formulated by
\begin{equation}\label{E}
	P_{h}^i=W_r^{3/2}\rho^{-1/2} sA^{-1/2}C_T^{-3/2}\frac{\delta}{8}+(1+k)\frac{W_r^{3/2}}{\sqrt{2\rho A}}.
\end{equation}

Thus, by summing the power consumption of all rotors, we can obtain the total hovering power consumption, denoted $P_{mh}$, as follows
\begin{equation}\label{F}
	\begin{aligned}
		P_{mh}=&\sum_{i=1}^{n}P_{h}^i\\
		=&n\times W_r^{3/2}\times\left(\rho^{-1/2} sA^{-1/2}C_T^{-3/2}\frac{\delta}{8}+\frac{(1+k)}{\sqrt{2\rho A}} \right)\\
		=&\underbrace{\frac{W^{3/2}}{\sqrt{n\rho A}}C_T^{-3/2}\frac{\delta}{8}s}_{\triangleq P_{bl}}+\underbrace{(1+k)\frac{W^{3/2}}{\sqrt{2n\rho A}}}_{\triangleq P_{in}},
	\end{aligned}
\end{equation} 
where $P_{bl}$ and $P_{in}$ denote the corresponding \emph{blade profile power} and \emph{induced power}, respectively. Combining \eqref{E} and \eqref{F}, $P_{h}^i$ and $P_{mh}$ can be approximated as two constants when UAVs hover at a fixed height. 
\begin{figure}[h]
	\centering
	\includegraphics[scale=0.39]{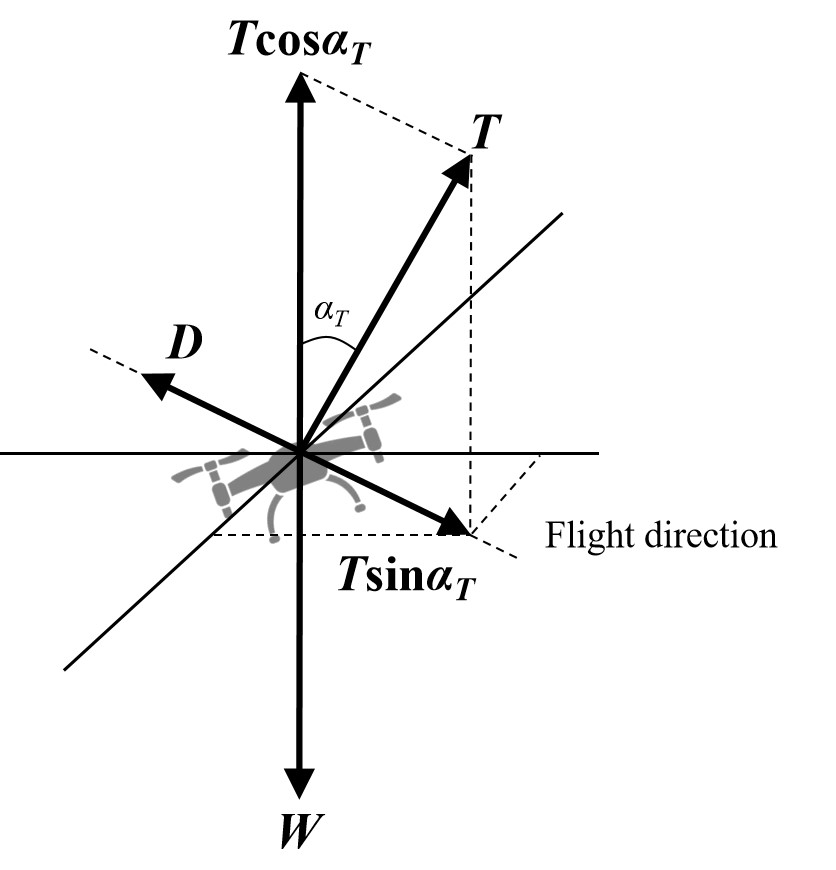}
	\caption{Schematic diagram of the forces acting on the  straightforwardly flying UAV at a fixed height.}
	\label{fig1}
\end{figure}

Given the multi-rotor UAV in forward flight, the simplified schematics of the forces acting on the  straightforwardly flying UAV at a fixed height is shown in Fig. \ref{fig1}, where $\alpha_T$ is the tilt angle of the rotor disc and the forces include rotor thrust $T$, fuselage drag $D$ and UAV weight $W$. Given a constant forward flight speed $V$, we can derive,
\begin{equation}\label{G}
\begin{aligned}
	T\cos\alpha_T=&W\\
	T\sin\alpha_T=&D.
\end{aligned}
\end{equation}
It follows from \eqref{G} that $T$ is equal to $\sqrt{W^2+D^2}$ based on \eqref{G}, but since $\alpha_T$ is usually very small when the UAV flys at a constant speed, we can have $T\approx W$, i.e., $\tilde{\kappa}\approx 1$. 

Clearly, the corresponding thrust-to-weight ratio (TWR) for the $i$-th rotor is also approximately 1. Furthermore, based on \eqref{B} and \eqref{E}, when the multi-rotor UAV flies forward (horizontally) with a constant speed $V$, the required power $P_{\mathrm{f}}^i(V)$ for the $i$-th rotor can be formulated as
\begin{equation}\label{I}
\begin{aligned}
P_{f}^i(V)=&W_r^{3/2} \rho ^{-1/2}sA^{-1/2}C_T^{-3/2}\frac{\delta}{8}
+\frac{3}{8}\delta \sqrt{\frac{W_r\rho A}{C_T}}sV^2 \\
+&(1+k)\frac{W_r^{3/2}}{\sqrt{2\rho A}}\left(\sqrt{1+\frac{V^4}{4v_0^4}-\frac{V^2}{2v_0^2}} \right)^{1/2}\\
+&\frac{1}{2}S_{FP\parallel}\rho V^3.
\end{aligned}
\end{equation}

Similarly, the corresponding required power $P_{mf}(V)$ for the multi-rotor UAV can be derived according to \eqref{I},
\begin{equation}\label{J}
	\begin{aligned}
		P_{mf}(V)=&\sum_{i=1}^{n}P_{f}^i(V)\\
		=&n\times \Bigg[ W_r^{3/2} \rho ^{-1/2}sA^{-1/2}C_T^{-3/2}\frac{\delta}{8}
		+\frac{3}{8}\delta \sqrt{\frac{W_r\rho A}{C_T}}sV^2 \\
		+&(1+k)\frac{W_r^{3/2}}{\sqrt{2\rho A}}\left(\sqrt{1+\frac{V^4}{4v_0^4}-\frac{V^2}{2v_0^2}} \right)^{1/2}\\
		+&\frac{1}{2}S_{FP\parallel}\rho V^3 \Bigg]\\
		=&P_{bl}+\frac{3}{8}\delta\sqrt{\frac{Wn\rho A}{C_T}}sV^2 
		+P_{in}\left(\sqrt{1+\frac{V^4}{4v_0^4}}-\frac{V^2}{2v_0^2} \right)^{1/2}\\
		+&\frac{n}{2}S_{FP\parallel}\rho V^3.
	\end{aligned}
\end{equation}

In conclusion, for a multi-rotor UAV, the power consumption models in hovering and forward flight, i.e. \eqref{F} and \eqref{J}, are respectively derived, and reflect the influence of the thrust, i.e. $T$. Similarly, the thrust will be further considered to obtain the power consumption models for a multi-rotor UAV in vertical flight in the following.
\subsection{Power Consumption Models for a Multi-Rotor UAV in Vertical Flight}
\begin{figure}[h]
	\centering
	\includegraphics[scale=0.42]{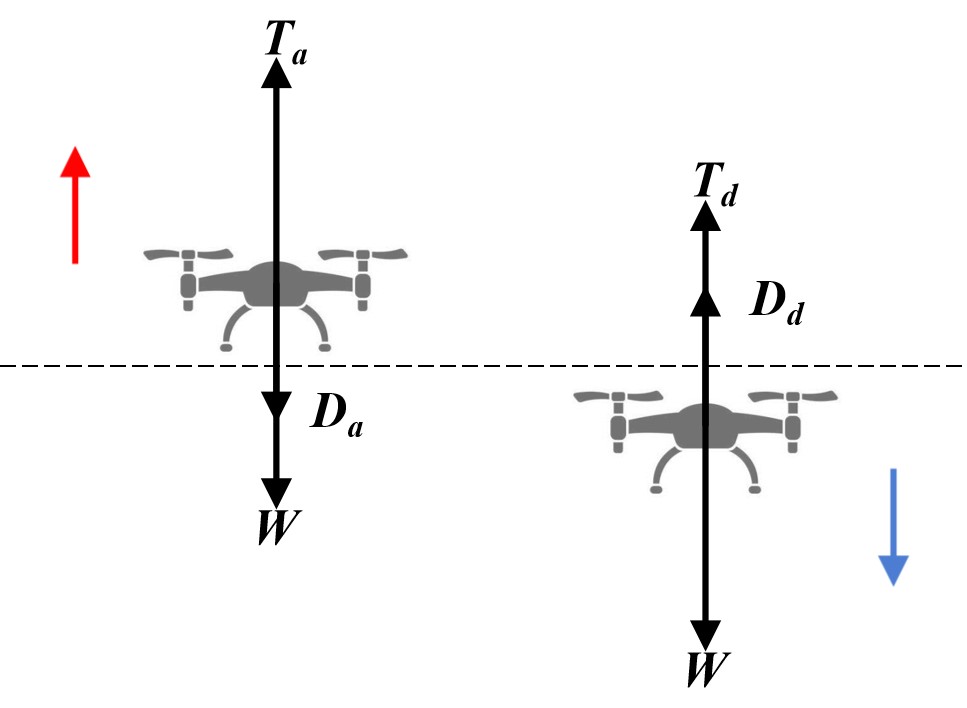}
	\caption{Schematic diagrams of the forces acting on the vertically flying UAV.}
	\label{fig2}
\end{figure}

According to the above results, the determination of the thrust is key to deriving the power consumption models for a multi-rotor UAV. Thus, we have to analyze the corresponding thrust for a multi-rotor UAV in vertical flight, including vertical ascent and vertical descent. Specifically, two schematic diagrams with respect to vertical ascent and vertical descent are illustrated in Fig. \ref{fig2} by involving the forces acting on the UAV,
where $T_a$ ($T_d$) and $D_{a}$ ($D_{d}$) denote the thrust and fuselage drag of the UAV associated with vertical ascent (descent), respectively.

Therefore, when the multi-rotor UAV is in vertical flight with a constant speed, we can have 
\begin{equation}\label{K}
	T_a-W=D_{a}
\end{equation}
for vertical ascent and 
\begin{equation}\label{K1}
	W-T_d=D_{d}
\end{equation}
for vertical descent.

When the multi-rotor UAV ascends or descends both at a constant speed, the corresponding thrusts $T_a$ and $T_d$ will be different, and in particular, since $D_a$ and $D_d$ are greater than 0, the former will be larger than the later regardless of the speed.

As discussed in the preceding subsection, the required power for the multi-rotor in vertical flight can be calculated by summing the power consumptions of all rotors. As such it is necessary to obtain the power consumption for each rotor by determining its thrust. Specifically, due to the equality among the thrust of each rotor given the multi-rotor UAV in vertical flight, the thrust of each rotor in vertical ascent is equal to $\frac{T_a}{n}$, i.e., $\frac{W+D_{a}}{n}$, and in vertical descent is equal to $\frac{T_d}{n}$, i.e., $\frac{W-D_{d}}{n}$. On these grounds, in combination with \eqref{K} and \eqref{K1}, the fuselage drags are equal to $\frac{D_{a}}{n}$ and $\frac{D_{d}}{n}$, respectively, and moreover, are often assumed to be $\frac{1}{2}S_{FP\perp}\rho V^2$ \cite{bramwell2001bramwell}. Note that $\rho$ is approximated to be constant since the changes in flight height have little effect on $\rho$.
Consequently, when the multi-rotor UAV is in vertical ascent or vertical descent, the corresponding thrust of the $i$-th rotor, denoted $T_{a}^i$ or $T_{d}^i$, can be further given by
\begin{equation}\label{L}
	\begin{aligned}
T_{a}^i=W_r+\frac{1}{2}S_{FP\perp}\rho V^2\\
T_{d}^i=W_r-\frac{1}{2}S_{FP\perp}\rho V^2.
	\end{aligned}
\end{equation}

Then, according to \eqref{C}, given the corresponding thrust in \eqref{L} and the speed $V$, the required power $P_{a}^i(V,T_{a}^i)$ and $P_{d}^i(V,T_{d}^i)$ for the $i$-th rotor in vertical ascent and vertical descent, respectively, can be expressed as
\begin{equation}\label{M}
	\begin{aligned}
P_{a}^i(V,T_{a}^i)=&P_{h}^i+\frac{1}{2}T_{a}^i V+\frac{T_{a}^i}{2}\sqrt{V^2+\frac{2T_{a}^i}{\rho A}},\\
P_{d}^i(V,T_{d}^i)=&P_{h}^i+\frac{1}{2}T_{d}^i V+\frac{T_{d}^i}{2}\sqrt{V^2+\frac{2T_{d}^i}{\rho A}}.
	\end{aligned}
\end{equation}

It is noticeable that the difference between $P_{a}^i(V,T_{a}^i)$ and $P_{d}^i(V,T_{d}^i)$ is caused by the thrusts in \eqref{L}, and this difference will become more obvious with V increasing.

According to \eqref{M}, the corresponding required power $P_{ma}(V,T_{a})$ and $P_{md}(V,T_{d})$ for the multi-rotor UAV can be calculated as
\begin{equation}\label{N}
	\begin{aligned}
		P_{ma}(V,T_{a})=&\sum_{i=1}^{n}P_{a}^i(V,T_{a}^i)\\
		=&n\times\left( P_{h}^i+\frac{1}{2} T_{a}^i V+\frac{T_{a}^i}{2} \sqrt{V^{2}+\frac{2 T_{a}^i}{\rho A}}\right)\\
		=&P_{mh}+\frac{1}{2}WV+\frac{n}{4}S_{FP\perp}\rho V^3\\
		+&\left( \frac{W}{2}+\frac{n}{4}S_{FP\perp}\rho V^2\right)\sqrt{(1+\frac{S_{FP\perp}}{A})V^2+\frac{2W}{n\rho A}}
	\end{aligned}
\end{equation}
and 
\begin{equation}\label{O}
	\begin{aligned}
		P_{md}(V,T_{d})=&\sum_{i=1}^{n}P_{d}^i(V,T_{d}^i)\\
		=&n\times\left( P_{h}^i+\frac{1}{2} T_{d}^i V+\frac{T_{d}^i}{2} \sqrt{V^{2}+\frac{2 T_{d}^i}{\rho A}}\right)\\
		=&P_{mh}+\frac{1}{2}WV-\frac{n}{4}S_{FP\perp}\rho V^3\\
		+&\left( \frac{W}{2}-\frac{n}{4}S_{FP\perp}\rho V^2\right)\sqrt{(1-\frac{S_{FP\perp}}{A})V^2+\frac{2W}{n\rho A}}.
	\end{aligned}
\end{equation}

Since $T_{a}$ and $T_{d}$ only vary with $V$ according to \eqref{L}, $P_{ma}(V,T_{a})$ and $P_{md}(V,T_{d})$ can be abbreviated as $P_{ma}(V)$ and $P_{md}(V)$, respectively. It follows that $P_{ma}(V)$ increases with $V$ increasing, but the change in $P_{md}(V)$ is difficult to be observed due to the complicated expression.
Besides, although both vertical ascent and vertical descent belong to the vertical flight of a multi-rotor UAV, the power consumption of the two is very different, and the former is theoretically larger, which will further affect the optimal path selection of the UAV.
\subsection{Generic Flight Power Model}
In previous subsections, the power consumption models for a multi-rotor UAV in horizontal flight and vertical flight have been obtained. However, these models are only considered in 1-D or 2-D scenarios. In practice, UAV usually flies in 3-D scenarios, and its flight status is not just horizontal or vertical flight, e.g. the flight status of rising and moving forward at the same time. 
Therefore, it is necessary to further study the power consumption for the UAV in a 3-D flight and establish the corresponding power consumption model for generic flight. 

\begin{figure}[h]
	\centering
		\includegraphics[scale=0.5]{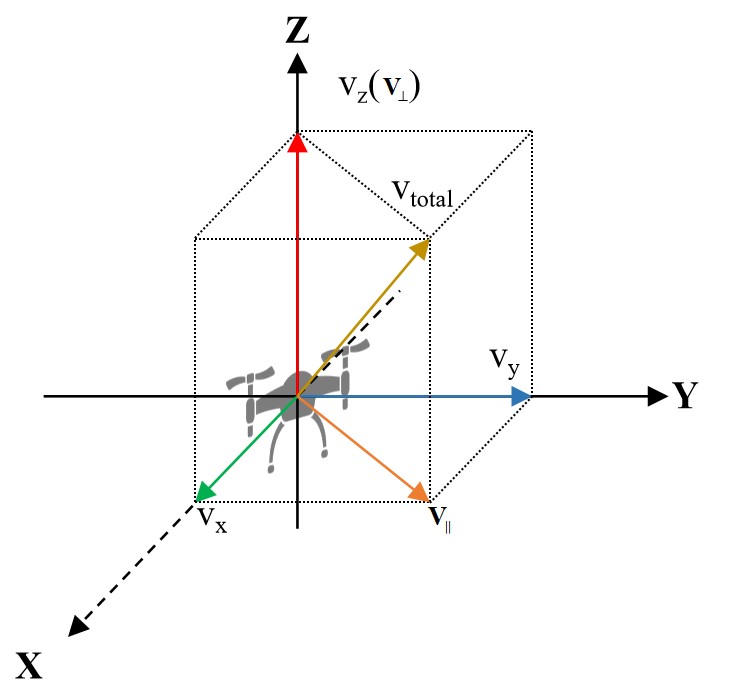}
	\caption{The UAV velocity in a 3-D scenario.}
	\label{fig6}
\end{figure}

Prior to modeling the generic power consumption model, the following factors should be known. First, velocity is the most critical factor in determining the power consumption according to the aforementioned. Second, the total power required for the multi-rotor UAV can be studied by analyzing the corresponding vertical and horizontal power consumption \cite{yan2021new}. 
A multi-rotor UAV flies in a 3-D scenario with the velocity $\mathbf{V}_{total}$, as shown in Fig. \ref{fig6}, which can be decomposed into three velocities in three distinct directions, namely $\mathbf{V}_\mathrm{x}$, $\mathbf{V}_\mathrm{y}$ and $\mathbf{V}_\mathrm{z}$ (or $\mathbf{V}_{\perp}$), respectively, with the three axes of X, Y, and Z representing the three directions of east, north and up, respectively. In addition, $\mathbf{V}_\parallel$ and $\mathbf{V}_\perp$ denote the horizontal velocity and vertical velocity of the UAV, respectively.

Evidently, $\mathbf{V}_{total}$ is determined by $\mathbf{V}_\parallel$ and $\mathbf{V}_\perp$, which reflect the UAV's flight status. 
Therefore, given $\mathbf{V}_{total}$, the total power consumption for the multi-rotor UAV in steady flight (i.e., without acceleration or deceleration) can be modeled as 
\begin{equation}\label{A1}
P_{total}(\mathbf{V}_{total})=P_{mh}+\Delta P_{\parallel}(\mathbf{V}_\parallel)+\Delta P_{\perp}(\mathbf{V}_\perp),
\end{equation} 
where $\Delta P_{\parallel}(\mathbf{V}_\parallel)$ is the power consumption increment in the horizontal direction with horizontal velocity $\mathbf{V}_\parallel$, and $\Delta P_{\perp}(\mathbf{V}_\perp)$ is the power consumption increment in the vertical direction with vertical velocity $\mathbf{V}_\perp$. 
Therefore, according to \eqref{J}, $\Delta P_{\parallel}(\mathbf{V}_\parallel)$ can be easily expressed as 
\begin{equation}\label{A2}
	\begin{aligned}
	\Delta P_{\parallel}(\mathbf{V}_\parallel)=&\frac{3}{8}\sqrt{n}\delta\sqrt{\frac{W\rho A}{C_T}}s \left\|\mathbf{V}_{\parallel}\right\|^2\\
	+&P_{in}\left[\left(\sqrt{1+\frac{\left\|\mathbf{V}_{\parallel}\right\|^4}{4v_0^4}}-\frac{\left\|\mathbf{V}_{\parallel}\right\|^2}{2v_0^2} \right)^{1/2}-1 \right]\\
	+&\frac{n}{2}S_{FP\parallel}\rho \left\|\mathbf{V}_{\parallel}\right\|^3,
	\end{aligned}
\end{equation}
where $\left\|\mathbf{V}_{\parallel}\right\|$ is the norm of the horizontal velocity $\mathbf{V}_\parallel$.

$ P_{\perp}(\mathbf{V}_\perp)$ involves $P_{ma}$ and $P_{md}$, and needs to be discussed in different situations due to the fact that
$P_{ma}$ and $P_{md}$ are the power consumptions in two different flight statuses, and do not hold when $\mathbf{V}_\parallel = 0$. Therefore, we have
\begin{equation}\label{A3}
P_{\perp}(\mathbf{V}_\perp)=\left\{
\begin{aligned}
	&P_{ma}(\left\|\mathbf{V}_{\perp}\right\|)      &\mathbf{V}_\perp > 0\\
	&P_{mh}                                          &\mathbf{V}_\perp = 0\\
	&P_{md}(\left\|\mathbf{V}_{\perp}\right\|)      &\mathbf{V}_\perp < 0\\
\end{aligned}
\right.,
\end{equation}
where
$\left\|\mathbf{V}_{\perp}\right\|$ is the norm of the vertical velocity $\mathbf{V}_\perp$. Besides, in order to ensure the integrity of \eqref{A3} and facilitate the following analysis, the case of $\mathbf{V}_\perp = 0$ is defined. 

Based on \eqref{N}, \eqref{O}, and \eqref{A3}, $\Delta P_{\perp}(\mathbf{V}_\perp)$ can be concisely formulated as
\begin{equation}\label{A4}
	\begin{aligned}
		\Delta P_{\perp}(\mathbf{V}_\perp)=&\frac{1}{2}W\left\|\mathbf{V}_{\perp}\right\|
		+sgn(\mathbf{V}_{\perp})\frac{n}{4}S_{FP\perp}\rho \left\|\mathbf{V}_{\perp}\right\|^3\\
		+&\left(\frac{W}{2}+sgn(\mathbf{V}_{\perp})\frac{n}{4}S_{FP\perp}\rho  \left\|\mathbf{V}_{\perp}\right\|^2\right)\\
		\times&\sqrt{(1+\frac{sgn(\mathbf{V}_{\perp})S_{FP\perp}}{A})\left\|\mathbf{V}_{\perp}\right\|^2+\frac{2W}{n\rho A}}\\
		+&(sgn(\left\|\mathbf{V}_{\perp}\right\|)-1)\frac{W}{2}\sqrt{\frac{2W}{n\rho A}} ,
	\end{aligned}
\end{equation}
where $sgn(\mathbf{V}_{\perp})$ is a sign function with respect to $\mathbf{V}_{\perp}$, which is used to distinguish the required power for vertical ascent and vertical descent in \eqref{A4}.  
\newcounter{TempEqCnt}
\setcounter{TempEqCnt}{\value{equation}}
\setcounter{equation}{18}
\begin{figure*}[ht]
	\hrulefill
\begin{equation}\label{B5}
\begin{aligned}
P_{total}(\mathbf{V}_{total})=&P_{\mathrm{mh}}+\frac{3}{8}\sqrt{n}\delta\sqrt{\frac{W\rho A}{C_T}}s \left\|\mathbf{V}_{\parallel}\right\|^2+P_{in}\left[\left(\sqrt{1+\frac{\left\|\mathbf{V}_{\parallel}\right\|^4}{4v_0^4}}-\frac{\left\|\mathbf{V}_{\parallel}\right\|^2}{2v_0^2} \right)^{1/2}-1 \right]
+\frac{n}{2}S_{FP\parallel}\rho \left\|\mathbf{V}_{\parallel}\right\|^3+\frac{1}{2}W\left\|\mathbf{V}_{\perp}\right\|\\
+&sgn(\mathbf{V}_{\perp})\frac{n}{4}S_{FP\perp}\rho \left\|\mathbf{V}_{\perp}\right\|^3
+\left(\frac{W}{2}+sgn(\mathbf{V}_{\perp})\frac{n}{4}S_{FP\perp}\rho  \left\|\mathbf{V}_{\perp}\right\|^2\right)\sqrt{(1+\frac{sgn(\mathbf{V}_{\perp})S_{FP\perp}}{A})\left\|\mathbf{V}_{\perp}\right\|^2+\frac{2W}{n\rho A}}\\
+&(sgn(\left\|\mathbf{V}_{\perp}\right\|)-1)\frac{W}{2}\sqrt{\frac{2W}{n\rho A}}
\end{aligned}
\end{equation}
\end{figure*}

As a result, when a multi-rotor UAV flies in a 3-D scenario with the velocity $\mathbf{V}_{total}$, $P_{total}(\mathbf{V}_{total})$ in \eqref{A1} can be further expressed by \eqref{B5} at the top of this page, which integrates the existing power consumption models for the multi-rotor UAV, and theoretically extends it to the 3-D scenarios, enriching the research in the field of UAV power consumption.

\section{Experiments and Simulations}
In this section, we first design three experiments to vaildate the above theoretical power models for a multi-rotor UAV in forward flight, vertical ascent and vertical descent, respectively. Then, we discuss the details of the corresponding data processing and show the experimental results. Finally, simulations are performed to investigate the influences of the rotor numbers on the power consumption for the UAV.
\begin{figure}[h]
\centering
\includegraphics[scale=0.5]{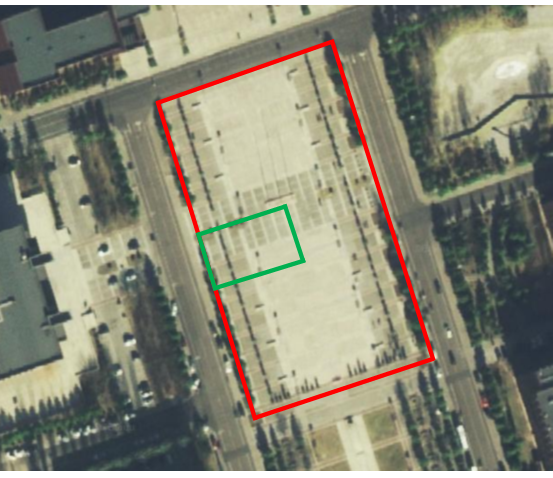}
\caption{The experiment fields. Experiment 1 is performed at the red area, while experiment 2 and 3 are at the green area.}
\label{fig3}
\end{figure}
\subsection{Experimental Setup}
\begin{figure}[h]
	\centering
	\includegraphics[scale=0.12]{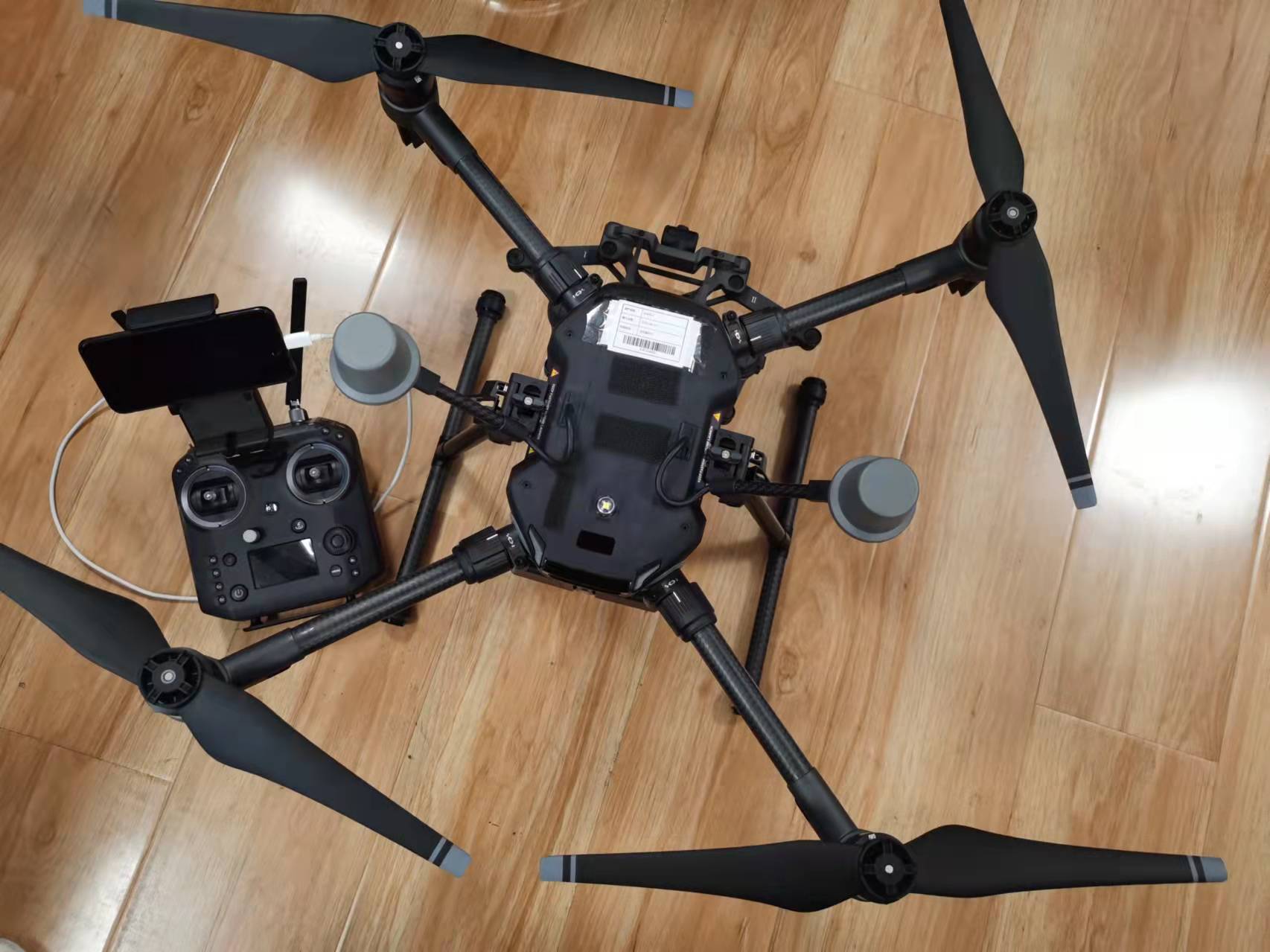}
	\caption{The UAV and its accessories used in the flight experiments.}
	\label{fig4}
\end{figure}

We have performed extensive flight experiments to actually collect the data for corresponding power consumptions for a multi-rotor UAV. To minimize the effect of wind and ensure obstacle-free flights, the experiments are performed in an open field with buildings on both sides. Fig. \ref{fig3} shows the field configuration and highlights the two experimental areas. Besides, all experiments are performed by using DJI M210 RTK V2 which is a quadrotor UAV. The actual UAV and its accessories used in the flight experiments as shown in Fig. \ref{fig4}. For all experiments, our intends are to vaildate the relationships between the flight speed and the power consumption, so we need to obatin different power consumption value at different flight speed. In flight experiments, the data such as the instantaneous flight speed, the instantaneous current and the instantaneous voltage of the UAV battery are recorded by our mobile app made by DJI Mobile SDK, with data collection frequencies of 1 Hz to speed and 10 Hz to current and voltage, i.e., one data measurement is obtained for every 1 seconds and 0.1 seconds, respectively.

For the intantaneous power consumption value, it can be obtained by multiplying the current flow by the voltage, which includes both the flying power and the communication-related power. According to our real-world flight experiments, the power for our UAV hover is around 300 W, while some wireless communication modules' power consumptions are about several hundreds of megawatts \cite{Lora2019,Esp2019}. Therefore, in the following experimental measurements, the communication-related power is ignored and the recorded power consumption for the UAV battery is treated as the flying power.  

\begin{figure}[h]
\centering
\includegraphics[scale=0.55]{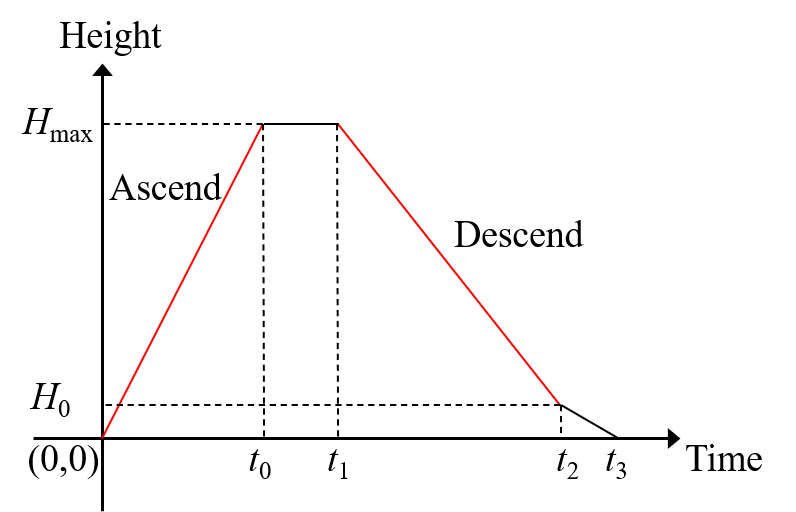}
\caption{The process of one round trip for the UAV in vertical flight.}
\label{fig5}
\end{figure}

In addition, we record the power consumptions for the UAV in steady flight at each specified speed and repeat each experiment for each targeting speed. Specifically, the first experiment focuses on the forward flight with an uniform speed, we let the UAV fly along a straight line with a fixed height 20 m, and we vary the speed from 0 m/s (hover) to 15 m/s with step 1 m/s. Due to the limitation of the field space to the straight flight, the UAV needs to fly several round trips to collect enough and balanced amount of the data for each specified speed. Similarly, for the second and third experiments which both focus on flying with an uniform speed in vertical status, we let the UAV ascend and descend at a constant vertical speed between 0 m and 110 m altitude as one round trip, where the ascent speed and descent speed are varied from 0 m/s to 5 m/s with step 0.5 m/s and 0 m/s to 3 m/s with step 0.5 m/s, respectively. Furthermore, the process of one round trip for the UAV in vertical flight is shown in Fig. \ref{fig5}, which we ignore the take off perparation time and the impact of acceleration and deceleration, the UAV first ascends with a constant speed to the max height $H_{max}$ in time duration $t_0$, then hovers for a short time from $t_0$ to $t_1$, and descends with a constant speed to height $H_0$ at $t_2$, which is the minimum safety height for continuous descent of the UAV, 
finally lands at $t_3$. So in above two experiments, we only record the data for the time period shown in the red area in Fig. \ref{fig5}.

\subsection{Data Processing}
Due to the different sampling rates for the UAV speed and the power consumption (related to current and voltage), one measurement of voltage and current corresponds to ten sets of speed data. Therefore, we use its average value to correspond to the one power consumption measurement by reason of the speed data collected within 1 seconds is close. Besides, different from experiment 2 and experiment 3, 
for each measurement in experiment 1, there are some data points obviously irrelevant to the specified speed, i.e., which is not equal to the specified speed value. This is because the larger speed range of the UAV in forward flight increases the difficulty of controlling the UAV (e.g., increases sensitivity of the control sticks) to maintain a steady flight. Denote by $P_i$ the measured power consumption, and the corresponding speed is $V_i$. In order to make full use of this data, we replace $V_i$ with $\lfloor {V}_{i}+\frac{1}{2}\rfloor$. 

\begin{figure*}
	\centering
	\subfigure[The UAV in forward flight.]{
		\begin{minipage}[t]{0.31\linewidth}
			\centering
			\includegraphics[scale=0.41]{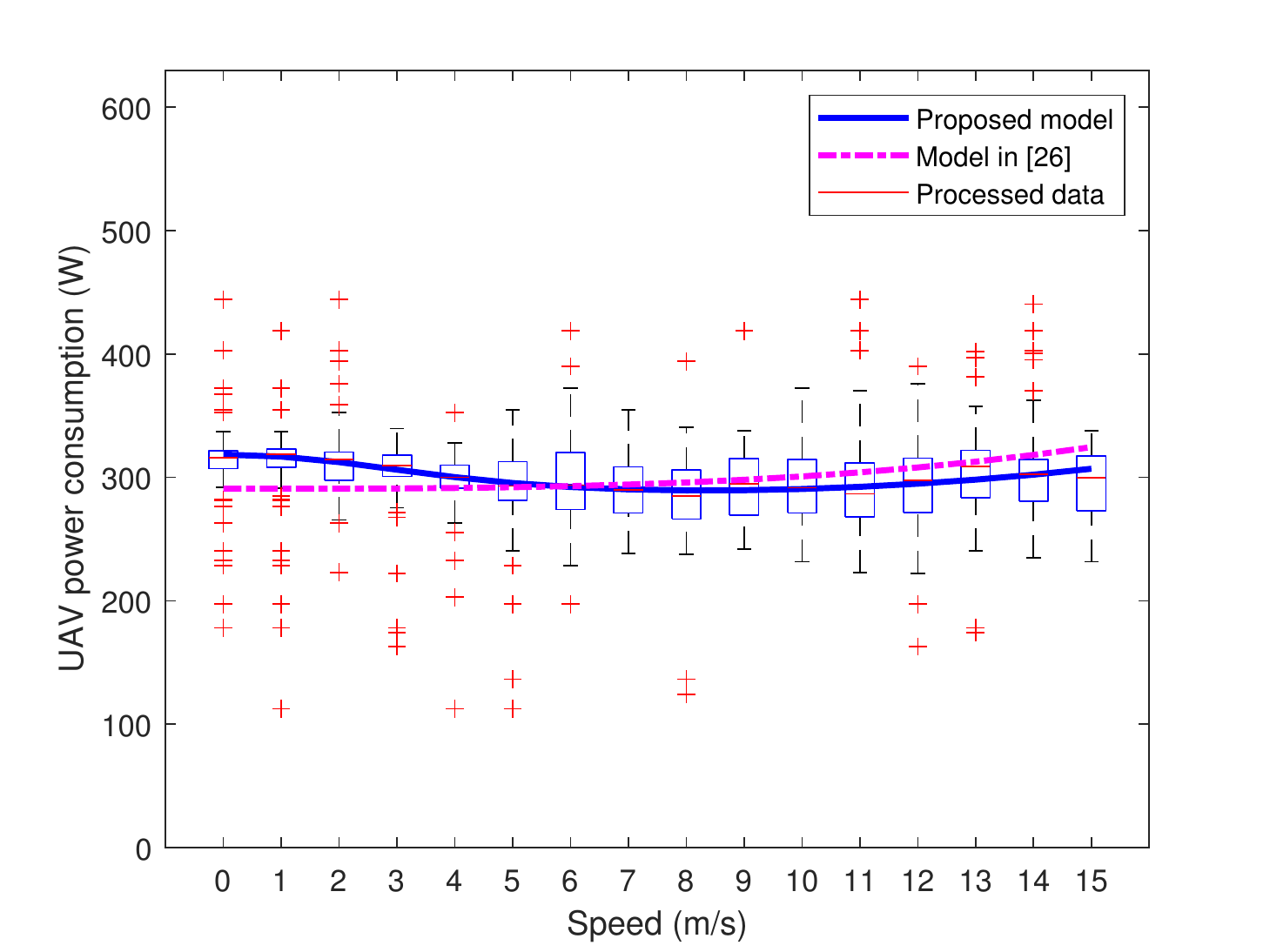}
			\label{fig11.1}
		\end{minipage}
	}
	\subfigure[The UAV in vertical ascent.]{
		\begin{minipage}[t]{0.31\linewidth}
			\centering
			\includegraphics[scale=0.41]{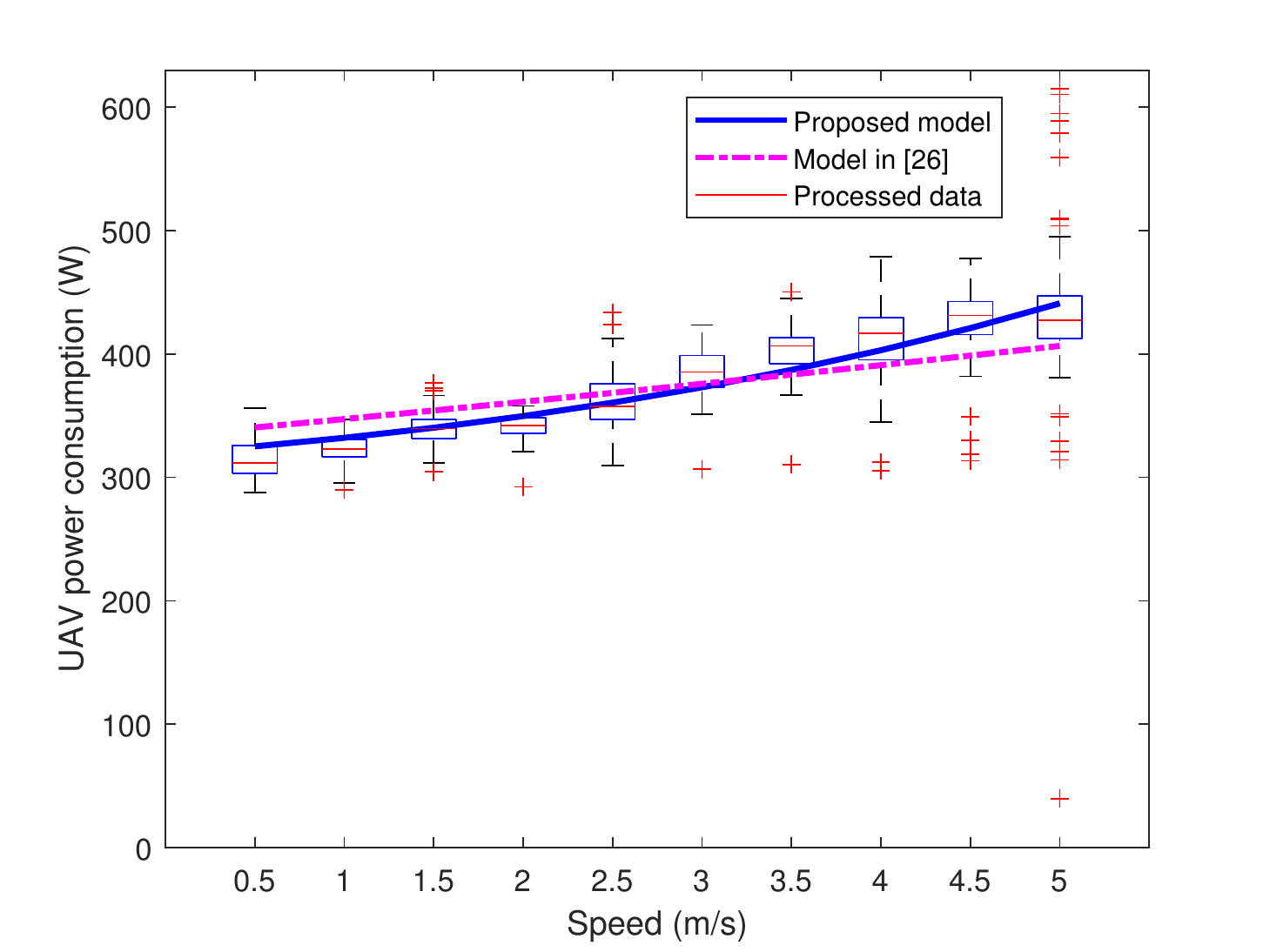}
			\label{fig11.2}
		\end{minipage}
	}
	\subfigure[The UAV in vertical descent.]{
		\begin{minipage}[t]{0.31\linewidth}
			\centering
			\includegraphics[scale=0.41]{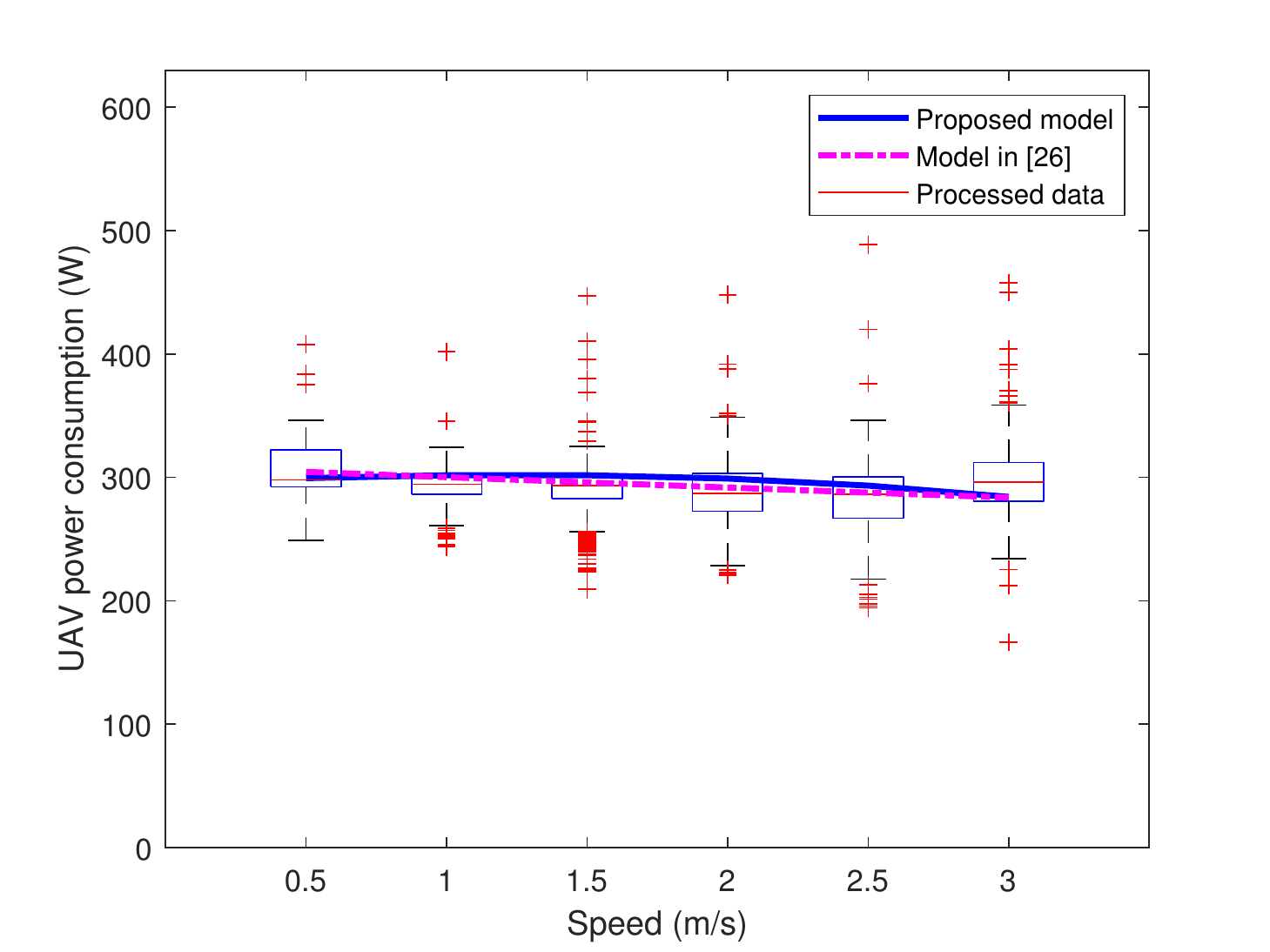}
			\label{fig11.3}
		\end{minipage}
	}
	
	\caption{The obtained power versus speed relationships with curve fitting based on proposed models and models [26].}
	\label{fig11}
\end{figure*}
\begin{figure*}
	\centering
	\subfigure[The UAV in forward flight.]{
		\begin{minipage}[t]{0.31\linewidth}
			\centering
			\includegraphics[scale=0.41]{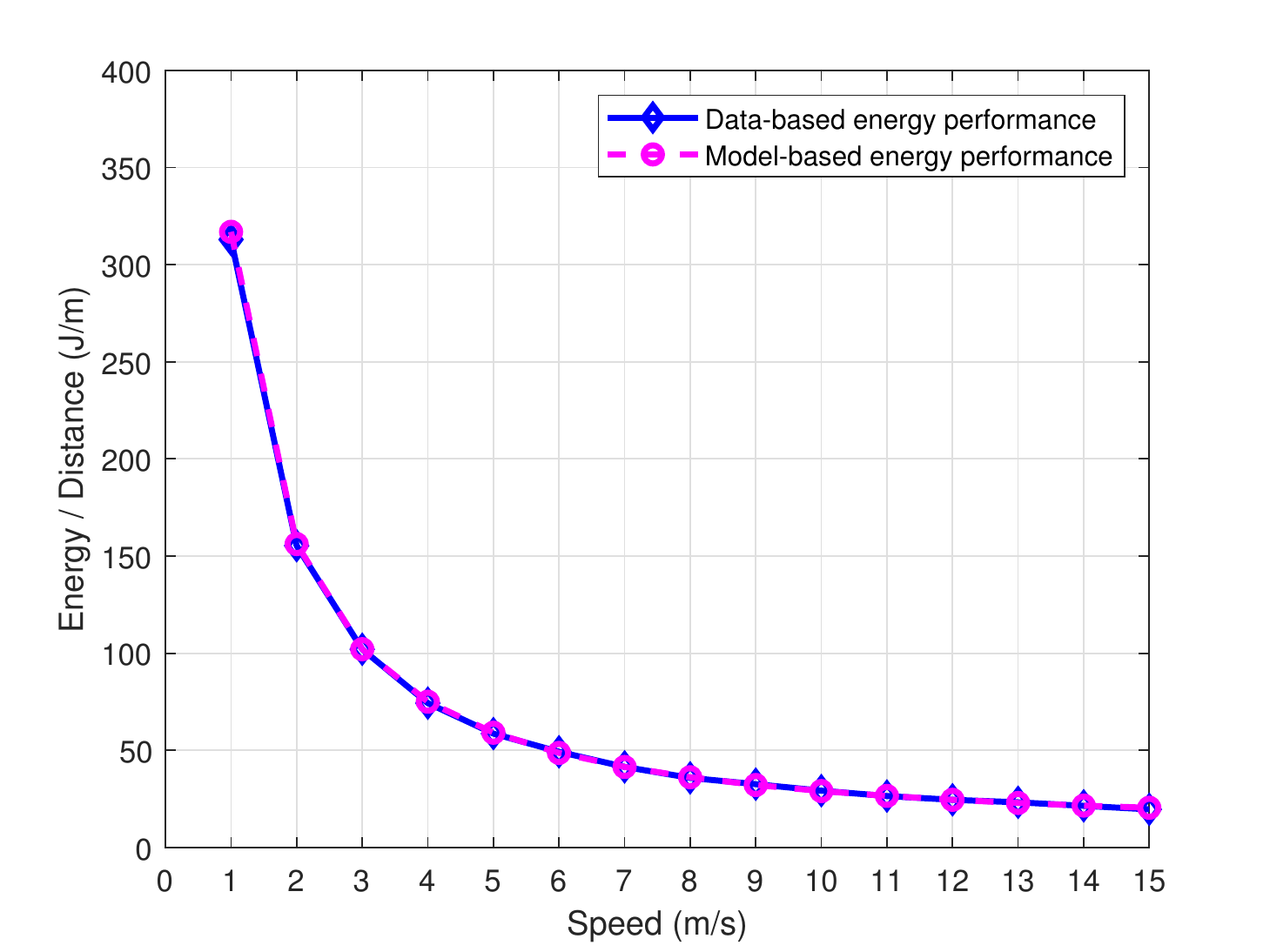}
			\label{fig9.1}
		\end{minipage}
	}
	\subfigure[The UAV in vertical ascent.]{
		\begin{minipage}[t]{0.31\linewidth}
			\centering
			\includegraphics[scale=0.41]{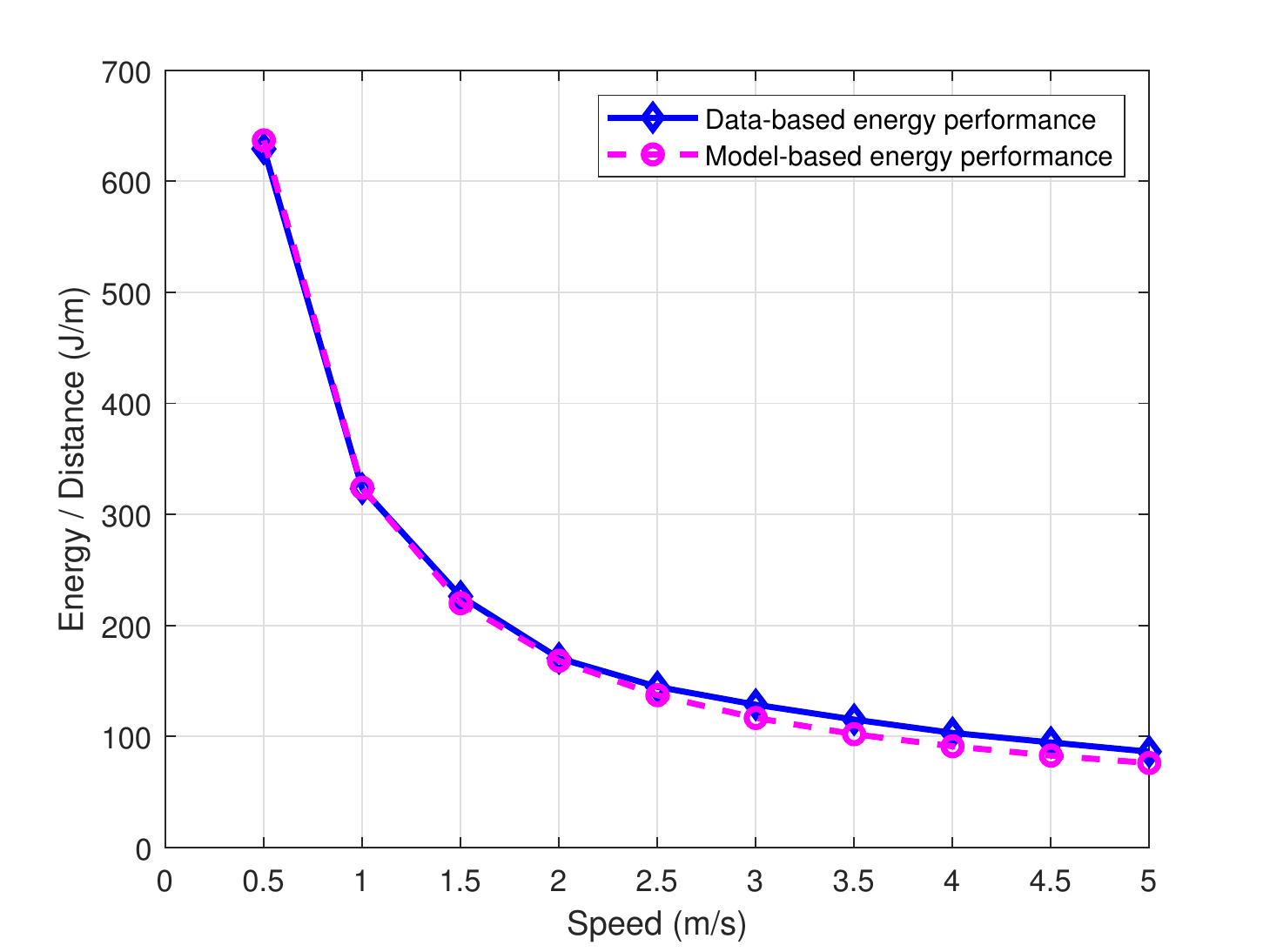}
			\label{fig9.2}
		\end{minipage}
	}
	\subfigure[The UAV in vertical descent.]{
		\begin{minipage}[t]{0.31\linewidth}
			\centering
			\includegraphics[scale=0.41]{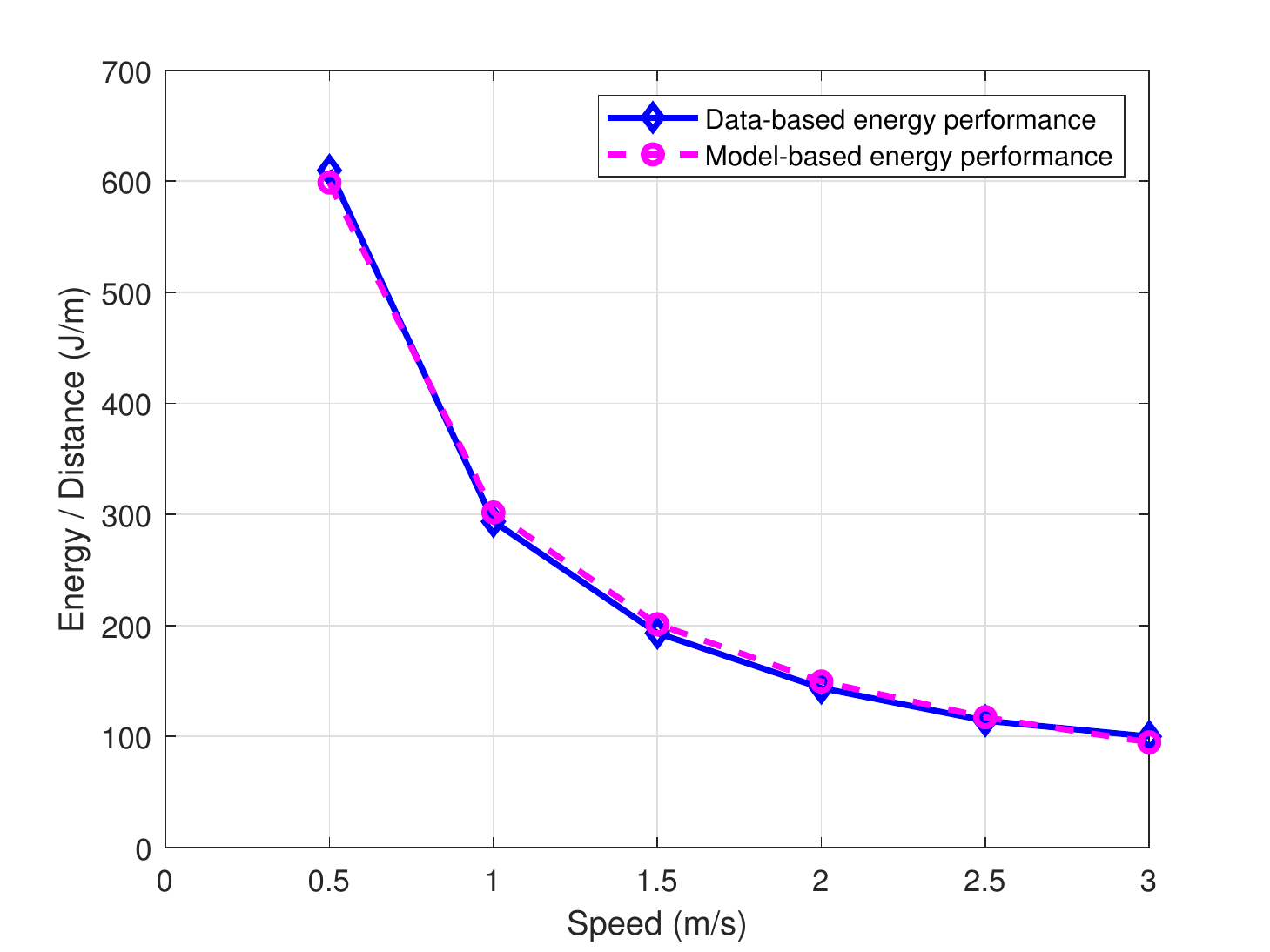}
			\label{fig9.3}
		\end{minipage}
	}
	
	\caption{The relationships between the UAV speed and the energy consumption per meter in three flight statuses.}
	\label{fig9}
\end{figure*}

Model-based curve fitting aims to fit data to user-defined model and has been applied to verify the power model \cite{gao2021energy}. Therefore, we use the experiment measurement data to fit our theoretical power consumption models to validate their correctness. However, the complexities of the proposed models lead to an infinite number of fitting results for some parameters. 
Thus, for better fitting results, we need to fit the power consumption models (i.e., \eqref{J}, \eqref{N} and \eqref{O}) in the form of combining parameters, can be respectively expressed as  
\begin{equation}\label{Q}
\begin{aligned}
P_{mf}^{\prime}(V) &=C_{1}+C_{2} V^{2}+C_{3}\left(\sqrt{1+\frac{V^{4}}{ C_{4}^{2}}}-\frac{V^{2}}{C_{4}}\right)^{1 / 2} \\
&+C_{5} V^{3},
\end{aligned}
\end{equation}
\begin{equation}\label{R}
\begin{aligned}
P_{ma}^{\prime}(V) &=C_{6}+C_{7} V+C_{8} V^{3} \\
&+\left(C_{7}+C_{8} V^{2}\right) \sqrt{\left(1+\frac{4 C_{8}}{C_{9}}\right) V^{2}+\frac{4 C_{7}}{C_{9}}},
\end{aligned}
\end{equation}
and
\begin{equation}\label{S}
\begin{aligned}
P_{md}^{\prime}(V) &=C_{6}+C_{7} V-C_{8} V^{3} \\
&+\left(C_{7}-C_{8} V^{2}\right) \sqrt{\left(1-\frac{4 C_{8}}{C_{9}}\right) V^{2}+\frac{4 C_{7}}{C_{9}}},
\end{aligned}
\end{equation}
where $C_i$, $i=1$, $\cdots$, $9$, are the modelling combination parameters.  

For all the measured data in experiment 1, denote the speed-power pair as ($V_{1}^i$, $P_{1}^i$) for the $i$-th data point after processing, similarly, denote by ($V_{2}^i$, $P_{2}^i$) and ($V_{3}^i$, $P_{3}^i$) the speed-power pairs in experiment 2 and experiment 3, respectively. With the least squares fitting, the purpose is to find suitable parameters $C_j$, $j=1$, $\cdots$, $9$, by minimizing the mean square error, i.e.,
\begin{equation}\label{U}
	\min _{C_{j}, j=1, \cdots, 5} \sum_{i=1}^{N_1}\left[P_{1}^i-P_{mf}'\left(V_{1}^i\right)\right]^{2},
\end{equation}
\begin{equation}\label{V}
	\min _{C_{j}, j=6, \cdots, 9} \sum_{i=1}^{N_2}\left[P_{2}^i-P_{ma}'\left(V_{2}^i\right)\right]^{2},
\end{equation}
and
\begin{equation}\label{W}
	\min _{C_{j}, j=6, \cdots, 9} \sum_{i=1}^{N_3}\left[P_{3}^i-P_{md}'\left(V_{3}^i\right)\right]^{2},
\end{equation}
where $N_k$, $k=1$, $\cdots$, $3$ denote the total number of the speed-power pairs in experiment 1, experiment 2 and experiment 3, respectively. Here, we use the built-in functionality of Matlab for curve fitting with the obtained data measurements.
\begin{figure*}
	\centering
	\subfigure[The UAV in forward flight.]{
		\begin{minipage}[t]{0.31\linewidth}
			\centering
			\includegraphics[scale=0.41]{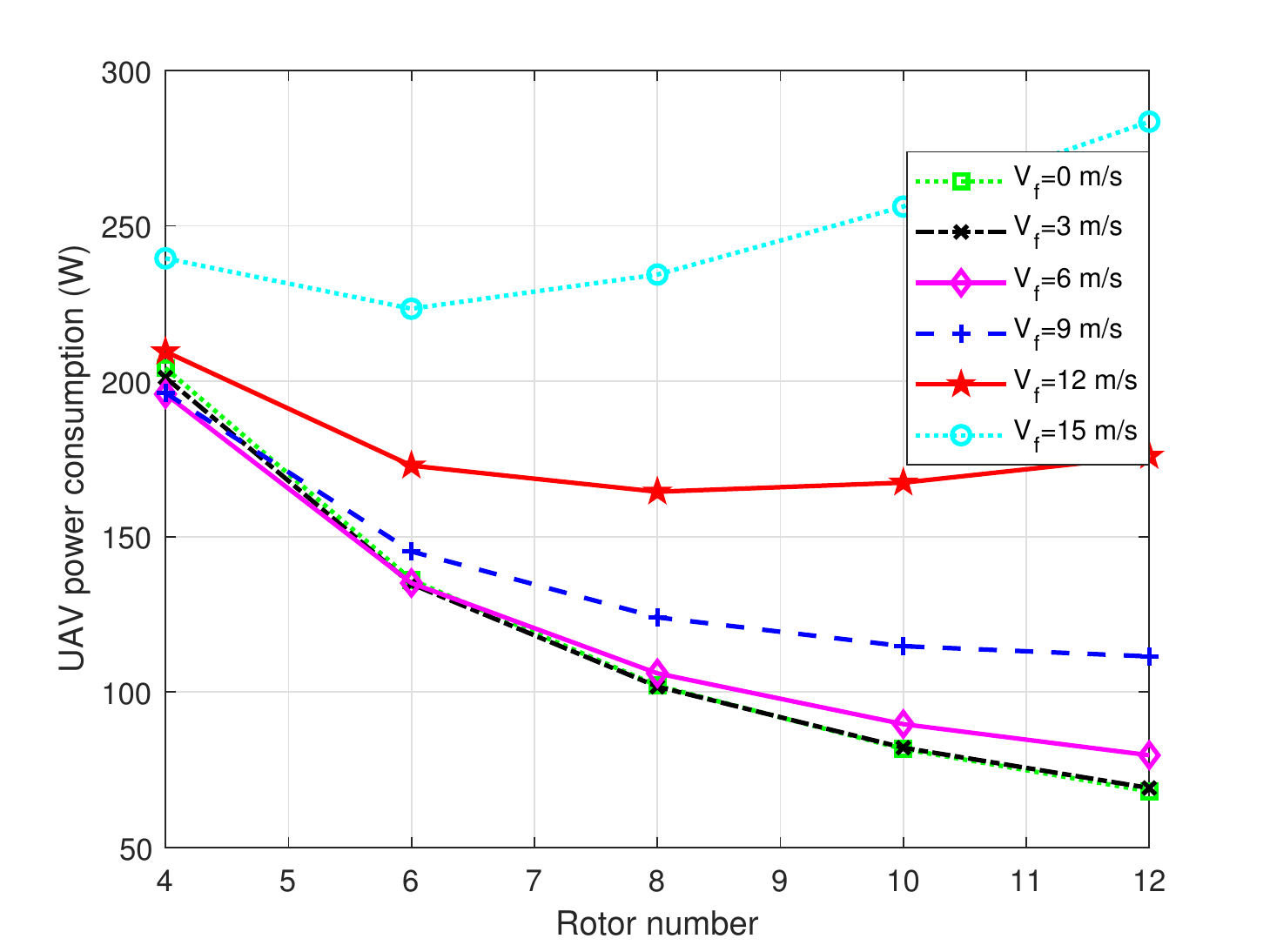}
			\label{fig8.1}
		\end{minipage}
	}
	\subfigure[The UAV in vertical ascent.]{
		\begin{minipage}[t]{0.31\linewidth}
			\centering
			\includegraphics[scale=0.41]{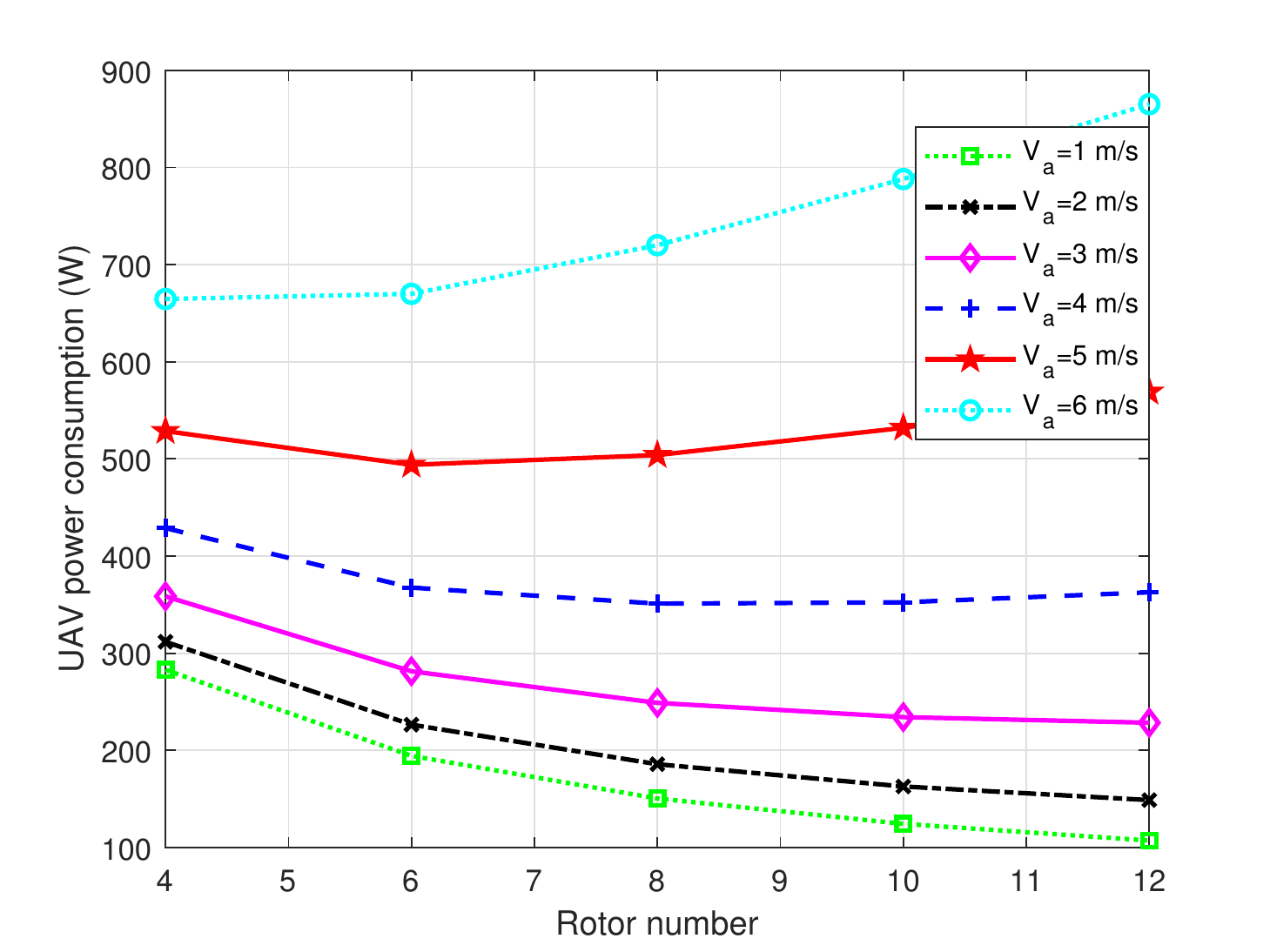}
			\label{fig8.2}
		\end{minipage}
	}
	\subfigure[The UAV in vertical descent.]{
		\begin{minipage}[t]{0.31\linewidth}
			\centering
			\includegraphics[scale=0.41]{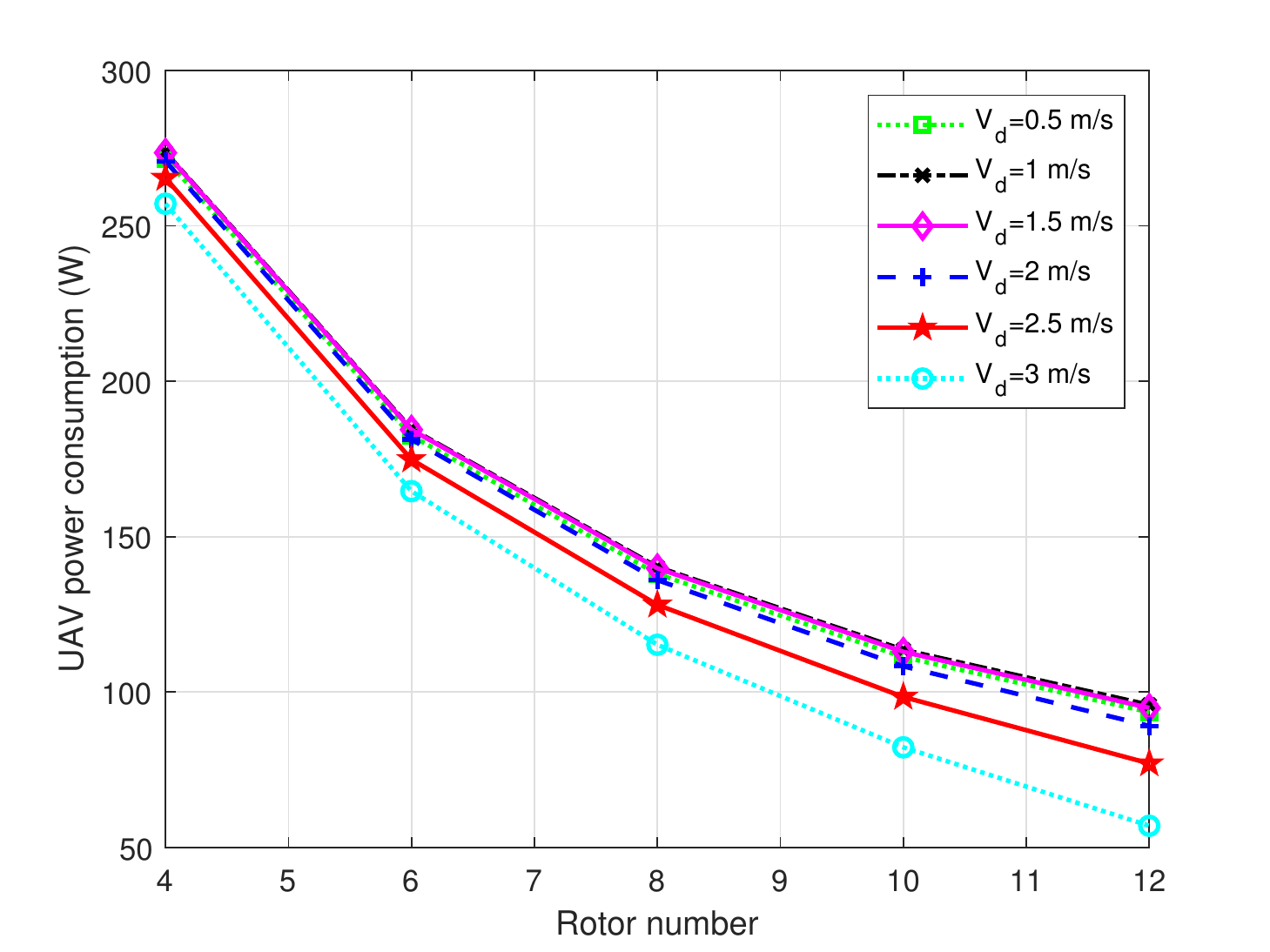}
			\label{fig8.3}
		\end{minipage}
	}
	
	\subfigure[The UAV in forward ascent.]{
		\begin{minipage}[t]{0.31\linewidth}
			\centering
			\includegraphics[scale=0.41]{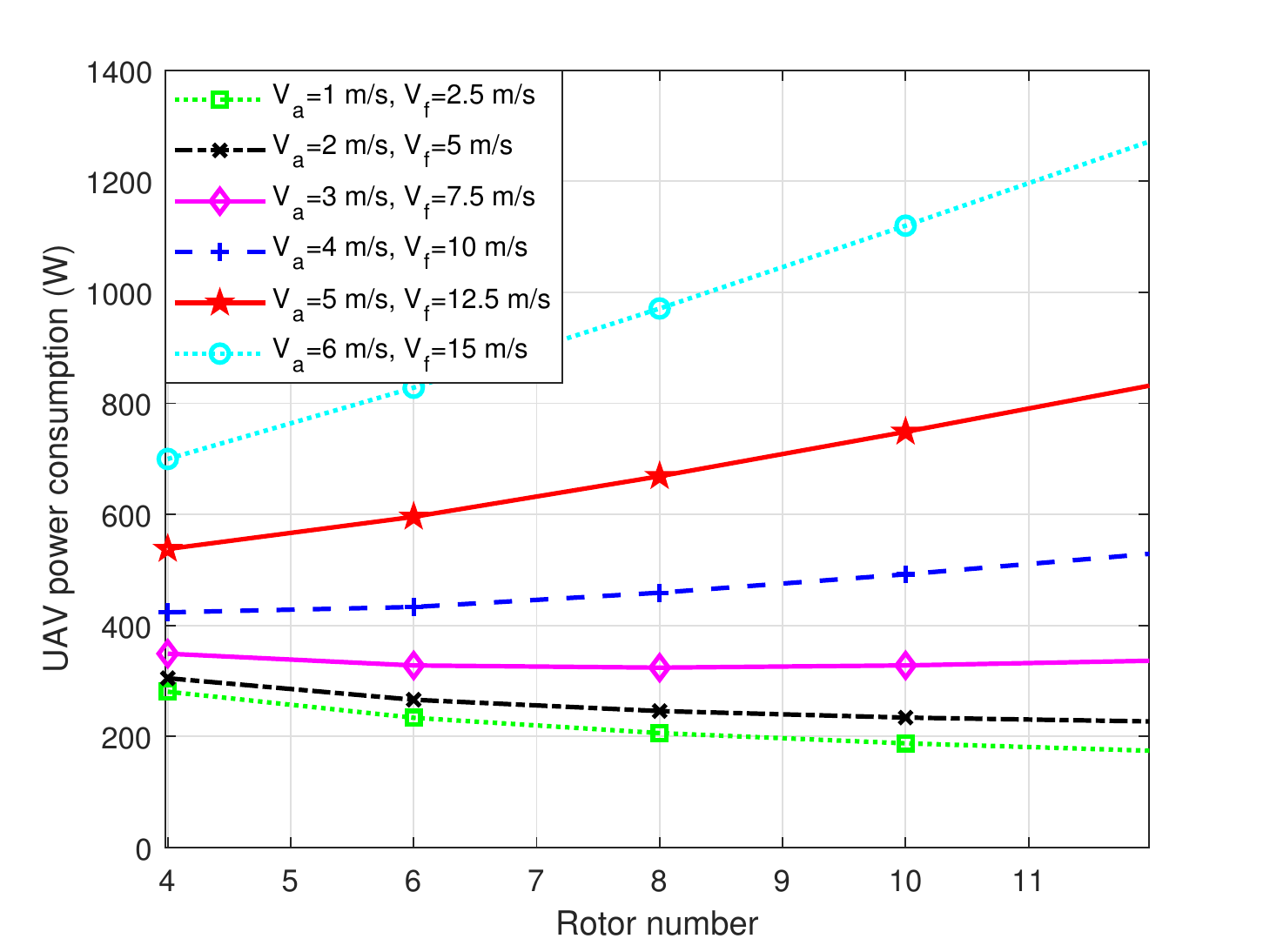}
			\label{fig8.4}
		\end{minipage}
	}
	\subfigure[The UAV in forward descent.]{
		\begin{minipage}[t]{0.31\linewidth}
			\centering
			\includegraphics[scale=0.41]{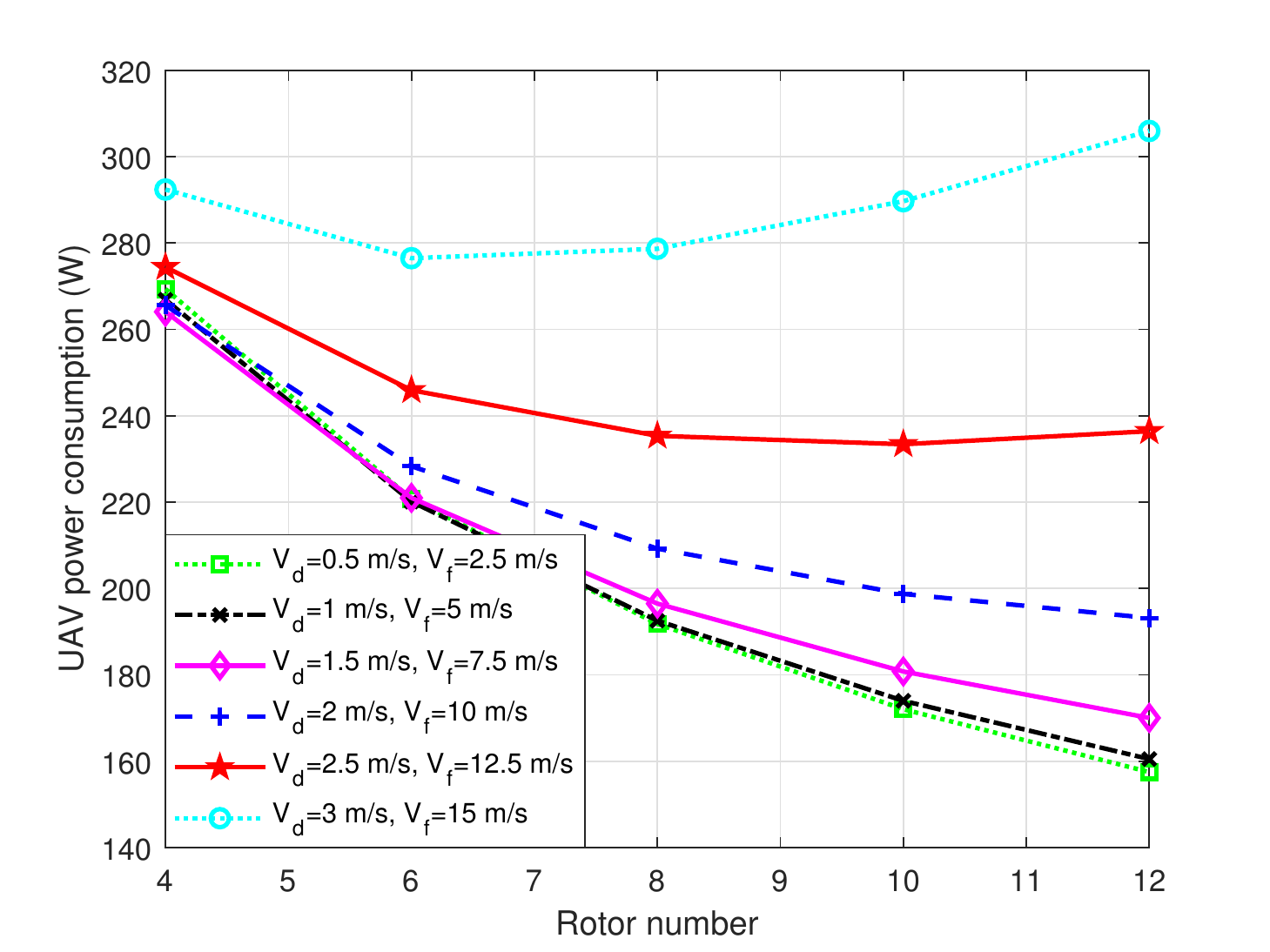}
			\label{fig8.5}
		\end{minipage}
	}
	\caption{The relationships between the rotor number and power consumption for the UAV in five flight statuses ($V_f$, $V_a$ and $V_d$ denote the forward flight speed, vertical ascent speed and vertical descent speed, respectively).}
	\label{fig8}
\end{figure*}
\subsection{Experimental Results and Discussions}
First of all, the processed data and the curve fitting results for the power consumption versus speed of the UAV are shown in Fig \ref{fig11}.
Specifically, it can be seen from the data expresed by box plots that there exists variations for the power even with the same speed; according to our best understanding, it is mainly caused by the wind, besides, there are also some other factors that can cause the power fluctuations, e.g., the temperature and the air pressure. The fitting curve results based on the theoretical power consumption models in \eqref{Q}, \eqref{R} and \eqref{S} are shown by the blue curves in Fig. \ref{fig11}. 
In order to further validate the accuracy of our proposed models, the fitted curves based on the models using aerodynamics in \cite{liu2017power} are shown by the pink curves; compare the two curves with the processed data, it is found that the proposed models fit the data better. Besides, the mean absolute error (MAE) and the root mean square error (RMSE) of the comparisons for the two curves with the data (i.e., the median of the power shown by box plots in Fig. \ref{fig11}) are summarized in Table \ref{tabl2}, which shows a good performance of our models. 
\begin{table}[h]
	\centering
	\begin{center}
		\caption{Results of Accuracy Comparison in Power Models}\label{tabl2}
		\begin{tabular}{|c|m{1.2cm}<{\centering}|m{1.2cm}<{\centering}|m{1.2cm}<{\centering}|m{1.2cm}<{\centering}|}
			\hline
			\textbf{Status} & \textbf{MAE (proposed model)} & \textbf{MAE (model in \cite{liu2017power} )} & \textbf{RMSE (proposed model)} & \textbf{RMSE (model in \cite{liu2017power} )}\\
			\hline
			\multirow{1}{*}{Forward flight } & 2.7296 & 13.4436 & 4.9228 & 16.4366\\
			\hline
			\multirow{1}{*}{Vertical flight} & 7.8554 & 12.1562 & 14.2425 & 15.4760\\
			\hline
		\end{tabular}
	\end{center}
\end{table}

Moreover, it is observed from Fig. \ref{fig11.1} that the fitting curve based on \eqref{Q} first decreases and then increases. With the speed increases, i.e., $V\geq11$ m/s, the power consumption increases significantly. Compared with the hover power (obtained by taking the average of the power measurements corresponding to the speed of 0 is about 317), it is found that the required power for forward flight is smaller in most situations. This phenomenon also appears in \cite{gao2021energy}, but without a further discussion. According to our analysis, this is because that when the status of the UAV changes from hovering to forward flight, the induced power becomes less. Although the parasite power and the blade profile power will increase with the speed increasing, the effect is small. 
In Fig. \ref{fig11.2}, the fitting curve based on \eqref{R} shows a noliner increase and 
the corresponding power for the multi-rotor UAV in vertical ascent is always higher than the hover power. In contrast, when the UAV is in vertical descent, the required power is less than that,
as shown by the blue curve based on \eqref{S} in Fig. \ref{fig11.3}. The is mainly because that the greater the thrust, the more the power required. As can be seen in Fig. \ref{fig11.3}, the steady blue curve indicates the effect of the speed on the UAV in vertical descent is small, hence the larger speed may be more suitable for the UAV's vertical descent in practice.


Additionally, since the energy for the UAV in actual flight is limited, it is of the great importance to operate the UAV at maximum performance. For the purpose of maximizing the total flight distance with any given energy, it is necessary to search the optimal speed \cite{hwang2018practical}. Hence we present the speed range in which the UAV (i.e., M210) can efficiently utilize energy based on the collected data, as shown by the blue lines in Fig. \ref{fig9}. In addition, due to the efficient utilization of energy can also be reflected by the power/speed ratio \cite{bramwell2001bramwell}, we obtain the ratio based on the proposed power consumption models, 
as illustrated by the pink dotted line in Fig. \ref{fig9}. Evidently, the two lines not only indicate the UAV energy consumption per unit flight distance in Joule/meter (J/m) with the speed, but also prove the validity of the models again. It is intuitive that the larger is the speed, the better is the energy utilization, but in theory there is a optimal speed (not the largest speed) when the UAV is in forward flight. However, this case is not implemented in this paper due to the speed exceeds the safe operating speed range of M210 (up to 15m/s without the wind). Therefore, we can determine the optimal method of operating the UAV by combining the optimum flight speed and the UAV's energy consumption along with the above mentioned results. 

In summary, the above extensive experiments validate the power consumption models for a multi-rotor UAV in forward flight, vertical ascent and vertical descent. Futhermore, the validation results support our analysis to more issues in the following. 
\subsection{Simulation Results}
This subsection provides simulation results to investiagte the effect of the rotor numbers on the power consumption for the UAV. The mainly parameters for the power consumption models are specified in Table \ref{tabl1} based on the above experiments, which are reasonable by comparing with \cite{zeng2019energy}. Besides, the thrust coefficient is $C_T = 0.001195$ and the fuselage equivalent flat plate area in vertical status is $S_{FP\perp} = 0.377$. The maximum flying speeds for the UAV in forward flight (horizontal flight), vertical ascent and vertical descent are 15 m/s, 6 m/s, and 3 m/s, respectively. Moreover, the rotor numbers for the UAV we considered is an even number between 4 and 12, where ten-rotor and twelve-rotor are two theoretical cases which is set to enrich and extend our simulation results. Note that the total rotor disc area is assumed unlimited for ease the simulation.
It is observed from Fig. \ref{fig8} that the change trend of the power consumption with the rotor numbers increasing is different and not always monotonic under different flight statuses and speeds for the UAV. Firstly, as can be seen from Fig. \ref{fig8.1} and Fig. \ref{fig8.2}, the UAV's power consumption gradually decreases with the rotor numbers increasing when $V_f$ ($V_a$) is not greater than 9 m/s (3 m/s), whereas, when the speed increases to 12 m/s (4 m/s), the corresponding trend of power consumption becomes first decrease and then increase. This is expected due to the following: the induced power is the main power consumption of the UAV while $V_f$ is small, and the induced drag will decrease with the rotor numbers increasing, so less power is needed; but when $V_f$ is large enough, more rotor numbers tend to lead a large increase in the parasite power, 
and therefore the required power becomes more. 

Moreover, as shown in Fig. \ref{fig8.3}, when the UAV is in vertical descent, its power consumpution decreases with the rotor numbers increasing. This is because although the increase in $V_f$ and the rotor number of the UAV will cause the parasite power to increase, unlike the situations in foward flight and vertical ascent, the parasite power here is actually used to propel the UAV, hence the actual power required by the UAV will become less. 

Additionally, in order to extend the investigation to more generic scenarios, according to \eqref{B5}, we consider the power consumptions for the UAV in two flight statuses i.e., the UAV ascends forwardly and descends forwardly with fixed directions (angles), besides, we set the ratio between $V_f$ and $V_a$ ($V_d$) as 2.5 (5) in the first (second) flight statuses. It is observed from Fig. \ref{fig8.4}, when the total speed (i.e., $\sqrt{V_a^2+V_f^2}$) is small, i.e., less than 8 m/s, more rotor numbers incure less power consumption, otherwise, more power consumtion is needed. When the total speed increases, the relationship between rotor numbers and power consumption tends to be linear. This phenomenon reflects the effect of the rotor numbers on the power consumption will decrease with the speed increasing. Besides, it is intuitive that the trend of the curves are similar in Fig. \ref{fig8.1} and Fig. \ref{fig8.5}, this is mainly because there is a large gap between the values of $V_f$ and $V_d$, the more power is used to maintain the horizontal flight.
  
In summary, although the influences of the rotor numbers on the power consumption for a multi-rotor UAV are different in different flight statuses, the same thing is that the power consumption decreases with the rotor numbers incresing. However, the influence is significant only when the UAV flight speed is small, with the speed increasing, the UAV's power consumption increases in most cases (except the UAV in vertical descent). 
\section{Conclusion}
This paper is in an effort to establish theoretical power consumption models for multi-rotor UAVs. To be specific, the power consumption models with closed-form expression for a multi-rotor UAV in three flight status, including forward flight, vertical ascent and vertical descent were established by analysing the relationship between single-rotor UAVs and multi-rotor UAVs in terms of power consumptions, and after that a generic flight power consumption model for the UAVs was obtained to satisify the power consumption expression for the UAV in more flight statuses. The extensive experiments confirmed the correctness of the models, and in order to investigate the influence of the rotor numbers on the power consumption of the UAV, the corresponding simulation was further conducted. 
\bibliographystyle{IEEEtran}
\bibliography{reference.bib}

\end{document}